\documentclass[Afour,sageh,times]{sagej}

\graphicspath{{./Figures/}{./Figures/Example_NoMove/}{./Figures/Example_Move/}{./Figures/Parametric_Analysis/}{./Figures/Parametric_Analysis/}{./Figures/Example_Graph/}{./Figures/Example_Coverage/}{./Figures/Experiments/}{./Figures/Example_Voronoi/}}
\DeclareGraphicsExtensions{.pdf,.jpeg,.png,.eps}

\usepackage{amssymb}
\usepackage{amsthm}
\usepackage{amsfonts}
\usepackage{mathtools}
\usepackage{algorithm}
\usepackage{algorithmic}
\usepackage{array}

\usepackage[caption=false,font=footnotesize]{subfig}

\usepackage{blindtext}
\usepackage{color}
\usepackage{comment}
\usepackage{balance}
\usepackage{enumerate}

\usepackage[outdir=./]{epstopdf}
\usepackage{soul}
\usepackage{multirow}
\usepackage[colorlinks,bookmarksopen,bookmarksnumbered,urlcolor=red,citecolor=black,linkcolor=black]{hyperref}
\usepackage{titlesec}
\titlelabel{\thetitle.\quad}

\newtheorem{theorem}{Theorem}[section]

\newtheorem{prop}[theorem]{Proposition}
\newtheorem{remark}[theorem]{Remark}

\newcommand{\blue}[1]{\color{blue}{#1}\color{black}\phantom{}}

\DeclareMathOperator*{\argmax}{argmax}

\newcommand\BibTeX{{\rmfamily B\kern-.05em \textsc{i\kern-.025em b}\kern-.08em
T\kern-.1667em\lower.7ex\hbox{E}\kern-.125emX}}

\begin{document}

\runninghead{Palacios-Gas\'os et al.}

\title{Equitable Persistent Coverage of Non-Convex Environments with Graph-Based Planning }

\author{Jos\'e~Manuel~Palacios-Gas\'os\affilnum{1}, Danilo~Tardioli\affilnum{2,3}, Eduardo~Montijano\affilnum{3} and Carlos~Sag\"u\'es\affilnum{3}}

\affiliation{\affilnum{1}Firmware Group, Induction Technology, Product Division Cookers, BSH Home Appliances Group, 50016, Zaragoza, Spain. \affilnum{2}Centro Universitario de la Defensa, Zaragoza, Spain. \affilnum{3}Instituto de Investigaci\'{o}n en Ingenier\'{\i}a de Arag\'{o}n (I3A), Universidad de Zaragoza, Zaragoza, Spain.}

\corrauth{Eduardo Montijano.}

\email{emonti@unizar.es}

\begin{abstract}
\textit{
In this paper we tackle the problem of persistently covering a complex non-convex environment with a team of robots. We consider scenarios where the coverage quality of the environment deteriorates with time, requiring to constantly revisit every point.
As a first step, our solution finds a partition of the environment where the amount of work for each robot, weighted by the importance of each point, is equal.
This is achieved using a power diagram and finding an equitable partition 
through a provably correct distributed control law on the power weights.
Compared to other existing partitioning methods, our solution considers a continuous environment formulation with non-convex obstacles.
In the second step, each robot computes a graph that gathers sweep-like paths and covers its entire partition.
At each planning time, the coverage error at the graph vertices is assigned as weights of the corresponding edges. Then, our solution is capable of efficiently finding the optimal open coverage path through the graph with respect to the coverage error per distance traversed.
Simulation and experimental results are presented to support our proposal.
}
\end{abstract}

\keywords{Persistent Coverage, Multi-Robot System, Equitable Partitioning, Non-Convex Environments, Path Planning}

\maketitle

\section{Introduction}
\label{sec:introduction}


Recent advances in mobile robotics and an increasing development of affordable autonomous mobile robots have motivated an extensive research in multi-robot systems. \footnote{\textcolor{red}{This is the accepted version of the manuscript: Palacios-Gasós JM, Tardioli D, Montijano E, Sagüés C. Equitable persistent coverage of non-convex environments with graph-based planning. The International Journal of Robotics Research. 2019;38(14):1674-1694. doi:10.1177/0278364919882082.
\textbf{Please cite the publisher's version}. For the publisher's version and full citation details see:\\
\protect\url{https://doi.org/10.1177/0278364919882082}. 
}}
The complexity of these systems resides in the design of communication, coordination and control strategies to perform complex tasks that a single robot is not able to do by itself.
A particularly interesting task is that of persistent coverage, in which a team of robots aims to maintain a desired coverage level over time in a given environment.
The particularity of persistent coverage, as opposed to other known coverage problems,~\cite{cortes2004coverage,Panagou:17}, is that the coverage level deteriorates with time. This implies that the robots need to revisit every point in the environment in order to keep the coverage at the desired levels.
This problem is of special interest in many applications such as vacuuming,~\cite{mackenzie1996vacuuming}, cleaning a place where dust is continuously settling,~\cite{kakalis2008oilspill},
lawn mowing,~\cite{arkin2000lawnmowing,sahin2007household} or environmental monitoring,~\cite{smith2011persistent,paley2008oceansampling}.
More recently, the apparition of useful unmanned aerial vehicles (UAVs) has encouraged the application of persistent coverage to surveillance and monitoring,~\cite{Renzaglia:12,nigam2014review}.

The solutions to this problem can be separated into two different groups, according to the division of the environment that is adopted to carry out the coverage mission.
In the first group of solutions, all the robots cover the entire environment together in a cooperative manner,~\cite{nigam2012uavsperssurv,atinc2014supervisedcov}, i.e., they are not restricted to particular areas.
The second group is composed of partition-based solutions. They address the problem with a divide-and-conquer strategy in the sense that they partition the environment and assign each partition to a single robot that becomes responsible of developing the coverage on it,~\cite{barrientos2011areacoveragepathplanning}.
The advantages of the second type are that, once the environment is partitioned, each robot can work independently, with no risk of collision and with no need for communication in the normal operation of the system.
For these reasons, we propose a solution of this type and pay special attention to the partitioning and the single-robot coverage inside each region.

\subsection{Environment Partitioning}

The most well-known partition technique for coverage in multi-robot systems is the Voronoi tesellation,~\cite{cortes2004coverage}.
Although different partition strategies for coverage had been previously proposed,~\cite{jager2002areapartcleaning}, the Voronoi partition has been the most used, especially for static coverage,~\cite{pierson2017adapting,moon2017voronoitracking}.
This type of partition has been successfully used in persistent coverage tasks for convex environments,~\cite{palacios2016control} or environments with few obstacles,~\cite{palacios2017pathplanning}. 
However, they present two important drawbacks.
In general, these partitions do not take into account the shape of the environment nor they account for obstacles and non-convexity.
The second and most important drawback is that, since they are purely geometric, the coverage share-out is not equitable.
This overloads some robots, while others have less work assigned than they can carry out.

Equitable partitions in convex environments are usually calculated in a centralized fashion. 
In~\cite{araujo2013uavdecompositioncov} a line in the optimal sweeping direction is moved along the environment to slice it each time the area on one of the sides is the same as any required partition.
A polygon decomposition is applied in~\cite{maza2007polygonareadecomp}, with the restriction of including the robot position inside its partition, and in~\cite{peters2017anytimeupdate} an algorithm handles sporadic communication between each robot and a central base to update the partition.
\cite{pavone2011distequitpartition} compute equitable partitions in convex environments using power diagrams in a distributed way.

On the other hand, the assumption of convex environments is not reasonable when it comes to ground vehicles.
In non-convex environments,
solutions based on Lloyd's algorithm appear in~\cite{pimenta2008herogeneous} and~\cite{breitenmoser2010voronoicovnonconvex}.
This algorithm is also combined with the geodesic distance in~\cite{Bhattacharya:14}, together with a graph-based path planning for exploration.
Frontier-based exploration of a non-convex environment is treated in~\cite{Hauman-discoverage11}.
Power diagrams are considered in~\cite{thanou2013distributed} for heterogeneous teams.
Complete coverage is guaranteed in~\cite{kapoutsis2017darp}, with a specifically tailored, optimality preserving, technique.
In~\cite{sea2017coordareapartition}, the features of online obstacle and decay learning are added to a decentralized partitioning of the target area on the basis of the robot performance.
The main difference of our approach with respect to these solutions is that the resulting partition is equitable in terms of the robots' workload.

Over discrete representations and graphs, Voronoi-based load-balancing partitions are presented in~\cite{boardman2016voronoiloadbalancing} and  in~\cite{durhan2012partition}, where it is combined with a gossip algorithm. Surfaces in 3D are treated in~\cite{breitenmoser2010distributed,breitenmoser2014adaptive}. Distributed vertex substitution is used in~\cite{yun2014graphequalmasspart}, with two-hop communication to guarantee convergence to the locally optimal configuration.
Compared to the aforementioned works, our partitioning approach works for a continuous environment formulation.

\subsection{Planning of Coverage Paths}

There are many possibilities to define the motion of the robot inside their partitions,~\cite{galceran2013covpathplanningsurvey,nigam2014review}. 
Some solutions posed the problem as a feedback control problem,
deciding at each time which is the best direction to move~\cite{franco2013persistent}. 
These solutions are easy to compute individually but typically lead the robots to local minima.
Our solution is also computationally light but considers a larger planning horizon.

A wide variety of planning approaches look for closed paths that each robot can follow periodically to cover its entire partition,~\cite{xu2014completecov}. 
Closed paths in square grids are computed in~\cite{arkin2000lawnmowing}.
Adaptive control is used by~\cite{soltero2013decentralized} and~\cite{lan2013planperstraject} consider Rapidly-exploring Random Cycles in Gaussian Random fields.
Discretized decay rates by a factor of 2 are considered in~\cite{smith2010multi} and~\cite{alamdari2014persistent} to compute different tours for several robots based on the importance of each location.
All these methods are closely related to the Traveling Salesman Problem (TSP). Their main advantage is that they typically allow for some measurable degree of optimality with respect to the optimal NP-hard solution. 
Compared to these methods, our planning algorithm considers fast online re-planning of open trajectories without being limited by the importance and decay factors of the vertices, at the cost of sacrificing bounds on the optimality gap with respect to the general problem.

Finally, we can find other similar solutions that deal with the online computation of open paths at particular times considering a finite-horizon.
A prioritized A* algorithm together with barrier functions is used in~\cite{Panagou:18}.
An integer program applied over a receding planning horizon is solved in~\cite{ahmadzadeh2007recedinghorizon} to find the paths that maximize spatio-temporal coverage.
In~\cite{graham2012adaptive} they obtain an approximate solution for a dynamic program over a finite time horizon, whose optimization criterion is the minimum uncertainty of a field estimate.
The local path planning algorithm \emph{TangentBug} is used in~\cite{breitenmoser2010voronoicovnonconvex} to plan the trajectories to the Voronoi centroids in a non-convex environment.
A line strip over a triangular mesh of equally important points is used in~\cite{breitenmoser2014adaptive}.
In~\cite{palacios2017pathplanning} a Fast Marching Method is used to compute optimal paths in terms of coverage quality to badly-covered areas.
The main difference of our solution with respect to these approaches is that we consider an optimization of the coverage cost per node visited.


\subsection{Contributions}

In this paper we propose a solution to persistently cover a complex, non-convex environment with a team of robots. Our solution works in two steps: the first one allows the robots to find an equitable partition of the environment in a distributed way, while the second one deals with the problem of individual online planning inside the partition using a graph representation.

The partitioning method builds upon the one presented in~\cite{pavone2011distequitpartition}. 
The main contribution is the extension of the method to a more general continuous environment formulation, including complex, non-convex ones, by developing the algorithm in terms of geodesic distance.
This formulation allows the partition to be consistent with the shape of the environment.
Additionally, we present two extensions of the algorithm to improve the partitions. The first one is designed to reduce the amount of disconnected partitions, in the sense that every partition is represented by a single connected component in the whole environment, avoiding the need to walk through other partitions to cover it.
The second extension allows the algorithm to compute equitable partitions weighted by the different capabilities of each robot.
In all cases, convergence is guaranteed.

The second part of our solution deals with the problem of individual path planning for persistent coverage inside each partition.
The method first generates a graph, discretizing the partition in a grid based on the coverage capabilities of the robot that will cover it.
Then, an online planner is run asynchronously by each robot, where at each planning time, the coverage error at the graph vertices is assigned as weights of the corresponding edges.
The planner finds then the path that maximizes the coverage quality per node visited using a variation of the Bellman-Ford algorithm (\cite{cormen2009algorithms}).
This allows the planning to be efficiently computed online with the particular advantages of using open paths that combined, resemble a sweeping strategy.

The remainder of the paper is structured as follows.
In Section~\ref{sec:formulation} we introduce the problem formulation.
In Section~\ref{sec:partition} we present the equitable partitioning method. 
We enhance the method to reduce disconnections in the partitions and consider energy constraints in Section~\ref{sec:connectivity}.
In Section~\ref{sec:coverage} the strategy to plan paths and individually cover each partition is presented.
Finally, we present simulation results in Section~\ref{sec:simulations}, experimental results in Section~\ref{sec:experiments} and conclusions in Section~\ref{sec:conclusions}.

\section{Problem Formulation}
\label{sec:formulation}

Let $\mathcal{Q} \subset \mathbb{R}^2$ be a known bounded environment, possibly non-convex, and $\mathcal{Q}_f\subseteq \mathcal{Q}$ the set of points in $\mathcal{Q}$ that are not obstacles.
The objective of persistent coverage in this environment is to maintain the actual coverage level, represented by a time-varying field, $Z(\mathbf{q},k) \geq 0$, as close as possible to a desired coverage objective, denoted as $Z^*(\mathbf{q}) > 0, \forall \, \mathbf{q} \in \mathcal{Q}_f$.
The importance of maintaining the coverage of each point is represented by a constant weighting function $\Phi(\mathbf{q}) \in (0,1]$.

The coverage level deteriorates over time with a constant decay rate, $d(\mathbf{q})$, with $0<d(\mathbf{q})<1$, according to the following recurrence equation:
\begin{equation}
Z(\mathbf{q},k) = d(\mathbf{q}) Z(\mathbf{q},k-1) + \alpha(\mathbf{q},k),
\label{Eq:VariationZ}
\end{equation}
where $\alpha(\mathbf{q},k)$ is the total coverage action that a team of robots applies at point $\mathbf{q}$ of the environment at time instant $k$ in order to satisfy the coverage objective.

The team is composed of $N$ robots, located at positions $\mathbf{p}_i(k), i \in \{1,\dots,N\}.$ 
The robots are capable of increasing the coverage in an area $\mathbf{\Omega}_i\big(\mathbf{p}_i(k)\big) \subseteq \mathcal{Q}_f$, that we call coverage area.
This area is bounded by a circle of radius $r_i^{cov}$, although it is not necessarily circular, convex or equal for all the robots.
We denote by $\alpha_i(\mathbf{q},k)$ the coverage value applied by the robot at position $\mathbf{q}$.
This value only depends on the physical properties of the robot, and outside of $\mathbf{\Omega}_i\big(\mathbf{p}_i(k)\big)$ is equal to zero.

In addition, we consider that the robots have the capability to adjust their production at each point of $\mathbf{\Omega}_i\big(\mathbf{p}_i(k)\big)$ by a coverage gain, $0\le \rho_i(\mathbf{q},k)\le 1.$ This way, the total coverage action is determined by
\begin{equation}
\alpha(\mathbf{q},k) = \sum_{i \in \{1,\dots,N\}} \rho_i(\mathbf{q},k) \, \alpha_i(\mathbf{q},k). 
\end{equation}
From now on, to simplify the notation we omit the spatial dependencies, $\mathbf{q},\ \mathbf{p}_i(k),$ in all the functions where it can be inferred by the context, e.g., $Z(\mathbf{q},k) \equiv Z(k), \alpha(k) \equiv \alpha(\mathbf{q}, k),\  \mathbf{\Omega}_i(\mathbf{p}_i(k))\equiv\mathbf{\Omega}_i(k),$ etc.

In the paper, we consider that the coverage gain is adjusted as in~\cite{palacios2017pathplanning}:
\begin{equation}
\rho_i(k) = \left\{
  \begin{aligned}
    & 1, & & \text{if } \rho_i^*(k) \geq 1, \\
    & \rho_i^*(k), & & \text{if } 0 < \rho_i^*(k) < 1, \\
    & 0, & & \text{if } \rho_i^*(k) \leq 0,
  \end{aligned}
  \right.
  \label{Eq:CovController}
\end{equation}
with 
\begin{equation}
\rho_i^*(k) = \frac{\displaystyle\int_{\mathbf{\Omega}_i(k)} \Phi\cdot \big( Z^* - d\; Z(k-1) \big) \alpha_i(k) \, \mathrm{d}\mathbf{q}}
{\displaystyle\int_{\mathbf{\Omega}_i(k)}\cdot \Phi \, \alpha_i(k)^2 \, \mathrm{d}\mathbf{q}}.
\end{equation}
Intuitively, $\rho_i^*(k)$ represents the fraction of the coverage action, $\alpha_i(k)$, required to reach the desired level, $Z^*$.
Therefore, what this action is doing is basically to adjust the coverage production to prevent coverage values above $Z^*$.

Now, we are interested in finding trajectories, $\Gamma_i(k)$, for all the robots that make $Z(k)$ as close as possible to $Z^*$ for the whole environment and every $k$.
Addressing this problem as a whole, e.g., as an optimization problem of the coverage error over some time horizon, is only possible for small teams of robots and very coarse discretized environments, as can be seen in~\cite{palacios2016branchandbound}.
The reason for this is that we need to account for the actions of multiple robots in the same environment, as well as the influence in the future coverage values of previous actions of the computed trajectories.
Therefore, in this paper we adopt a divide and conquer strategy to address the problem in two stages.

In a first stage, described in Sections~\ref{sec:partition} and~\ref{sec:connectivity}, we focus on dividing the environment into $N$ disjoint regions, one for each robot, such that the work needed to cover them, in terms of coverage action is equal. This way, we simplify the problem of finding $N$ paths into $N$ individual problems of finding a single path in a smaller environment. This partition can be computed once offline and be assumed constant afterwards.

In a second stage, detailed in Section~\ref{sec:coverage}, we propose an online planning algorithm that each robot executes locally within its partition to obtain the trajectories $\Gamma_i(k)$ that minimize the \emph{current} coverage error. The most important simplification we make to keep the problem tractable is to disregard in the planning algorithm the influence of previous actions of the computed trajectory into the coverage value.

\section{Equitable Partitioning}
\label{sec:partition}

The first step of our strategy to solve the persistent coverage problem is to divide the environment in as many equitable regions as robots. The goal is that each partition requires an amount of work proportional to the capabilities of its assigned robot with respect to the rest of the team and to the importance of its points.
To this end, we introduce the notions of power diagram and geodesic distance, define the \mbox{importance-weighted} workload of the agents in terms of the coverage problem and detail our distributed partitioning algorithms. 

\subsection{Power Diagrams and Geodesic Distance}

Let us first define a set of points, $\mathcal{G} = \{\mathbf{g}_1, \dots, \mathbf{g}_N\}$, with $\mathbf{g}_i, \mathbf{g}_j \in \mathcal{Q}_f$ for all $i$ and $j$ and $\mathbf{g}_i \neq \mathbf{g}_j$ if $i \neq j$, which will act as the generators of the partitions.
Our algorithm to find the equitable partitions is based on power diagrams~\cite{aurenhammer1987power}, that are a generalization of the Voronoi diagrams.
Instead of assigning the points to each partition according to the Euclidean distance to the generator, the power diagrams use the squared distance minus a certain weight. If we let $\mathbf{w} = \{w_1,\dots,w_N\}$ be the set of power weights associated to the partitions, we can formally define the partitions obtained from the power diagram as
\begin{equation}
    \mathcal{P}_i(\mathbf{w}) = \{ \mathbf{q} \in \mathcal{Q}_f \,  | \, d(\mathbf{q},\mathbf{g}_i)^2 - w_i \leq d(\mathbf{q},\mathbf{g}_j)^2 - w_j \}.
    \label{Eq:PowerDiagram}
\end{equation}

In order to apply this partitioning to a non-convex environment, we make use of the geodesic distance, $d_g(\mathbf{q},\mathbf{g})$, the length of the shortest path between two points of the environment.
Assuming that the environment is composed of polygonal obstacles, the shortest path between every two points is the concatenation of a set of straight lines connecting the two points and some vertices of the obstacles, $\{\mathbf{q}, \mathbf{h}_{\mathbf{q},\mathbf{g}}^1, \mathbf{h}_{\mathbf{q},\mathbf{g}}^2, \dots, \mathbf{h}_{\mathbf{q},\mathbf{g}}^{l_{\mathbf{q},\mathbf{g}}}, \mathbf{g}\}$, where $l_{\mathbf{q},\mathbf{g}}$ is the number of obstacle vertices that define such path.
Therefore, the geodesic distance can be decomposed in the summation of the Euclidean distances between each pair of consecutive points:
\begin{align}
    d_g(\mathbf{q},\mathbf{g}) = & \| \mathbf{q} - \mathbf{h}_{\mathbf{q},\mathbf{g}}^1 \| + \| \mathbf{h}_{\mathbf{q},\mathbf{g}}^1 - \mathbf{h}_{\mathbf{q},\mathbf{g}}^2 \| \nonumber \\
    & + \ldots + \| \mathbf{h}_{\mathbf{q},\mathbf{g}}^{l_{\mathbf{q},\mathbf{g}}} - \mathbf{g} \|.
    \label{Eq:GeodDistDecomposition}
\end{align}

Using this metric, the boundary between two partitions is
\begin{equation}
    \Delta_{ij} = \{ \mathbf{q} \in \mathcal{Q}_f \, | \, d_g(\mathbf{q},\mathbf{g}_i)^2 - w_i = d_g(\mathbf{q},\mathbf{g}_j)^2 - w_j \}.
    \label{Eq:BoundaryPoints}
\end{equation}
According to this definition, the neighbors of a generator are
\begin{equation}
    N_i = \{ j \in \{1,\dots,N\} \setminus i \ | \, \Delta_{ij} \neq \emptyset \}.
\end{equation}
We assume that each robot can communicate with robots of neighboring regions.

\subsection{Importance-Weighted Workload}

The partitions that are assigned to the robots have to be equitable in terms of the work that has to be carried out inside them.
At the same time, they have to take into account the importance of the coverage of each point. Therefore, it is essential to define the workload in terms problem's variables.
The definition of the work at each point of the environment that we consider is
\begin{equation}
\label{Eq:importanceFunc}
\lambda(\mathbf{q}) = \Phi \, ( 1 - d ) \, Z^*.
\end{equation}
Strictly speaking, $( 1 - d ) \, Z^*$ is the actual workload of each each point. It represents the coverage that decays at each time if the point has reached the desired level, i.e., in the steady state.
Nevertheless, we weight it with the importance to allow the robots to spend more time covering the most important points, at the expense of reducing the attention to less important points.
This may lead to partitions in which $( 1 - d ) \, Z^*$ is not equitable but the importance of the work of each partition is.
From now on, we refer to this importance-weighted workload simply as workload.

The workload inside each partition can be calculated as
\begin{equation}
\lambda_{\mathcal{P}_i(\mathbf{w})} = \int_{\mathcal{P}_i(\mathbf{w})} \lambda(\mathbf{q}) \, d\mathbf{q},
\end{equation}
Both the importance and the decay take values between $0$ and $1$. On the contrary, the desired coverage level is expressed in general in coverage units and, therefore, we normalize the entire workload by $\int_{\mathcal{Q}_f} \Phi \, ( 1 - d ) \, Z^* \, d\mathbf{q}$ to obtain it in percentages.

\subsection{Distributed Algorithm for Equitable Partitioning}

The power diagram can be seen as a relation between the weights and the regions that belong to their corresponding generators.
Therefore, we minimize a cost function whose minima correspond to sets of weights that lead to an equitable partition:
\begin{equation}
\label{Eq:CostfunctionPartition}
    H(\mathbf{w}) = \sum_{i=1}^N \frac{1}{\lambda_{\mathcal{P}_i(\mathbf{w})}}.
\end{equation}
One can see that this function reaches its minimum when the workload of all the partitions is the same.
It is also important to remark that such minimum can always be found, see~\cite{pavone2011distequitpartition}.
From now on, we omit the dependencies of $H$ and $\mathcal{P}_i$ with $\mathbf{w}$ for the sake of clarity.

To minimize this function distributively, each robot is in charge of a generator, that we locate at the robot's position for simplicity, and its associated weight, initially set to zero.
The evolution of the weight is driven to achieve equity using information from the neighbors in the following control law.


\begin{theorem}
    The power diagram generated by $\mathcal{G}$ and $\mathbf{w}$ converges to an equitable power diagram under the distributed control law
    \begin{equation}
        \dot{w}_i = - k_w \,\frac{\partial H}{\partial w_i},
        \label{Eq:ControlLaw1}
    \end{equation}
    with $k_w$, a positive gain, and
\begin{equation}
    \frac{\partial H}{\partial w_i} = \sum_{j \in N_i} \left( \frac{1}{\lambda^2_{\mathcal{P}_j}} - \frac{1}{\lambda^2_{\mathcal{P}_i}} \right) \int_{\Delta_{ij}} \dfrac{\lambda(\mathbf{q})}{\| n_{ij}'(\mathbf{q}) \|} \, d\mathbf{q},
    \label{Eq:dH_dwi}
\end{equation}
where $n_{ij}'(\mathbf{q})$ is the outward normal to the boundary between the partitions $i$ and $j$ at point $\mathbf{q} \in \Delta_{ij}$.
\label{Theo:Equitable}
\end{theorem}
\begin{proof}
In the first place we develop the gradient of the cost function with respect to a single weight $w_i$: 
\begin{equation}
    \frac{\partial H}{\partial w_i} = - \frac{1}{\lambda^2_{\mathcal{P}_i}} \frac{\partial \lambda_{\mathcal{P}_i}}{\partial w_i} - \sum_{j \in N_i} \frac{1}{\lambda^2_{\mathcal{P}_j}} \frac{\partial \lambda_{\mathcal{P}_j}}{\partial w_i}.
    \label{Eq:dH_dwi_0}
\end{equation}
It only depends on the neighboring partitions since a variation on $w_i$ only affects them.
The derivative of the workload in the partition of robot $i$ is
\begin{equation}
    \frac{\partial \lambda_{\mathcal{P}_i}}{\partial w_i} = \frac{\partial}{\partial w_i} \int_{\mathcal{P}_i} \lambda(\mathbf{q}) \, d\mathbf{q}.
\end{equation}
It can be transformed using the following result associated with the divergence theorem:
\begin{equation}
    \frac{\partial}{\partial w_i} \int_{\mathcal{P}_i} \lambda(\mathbf{q}) \, d\mathbf{q} = \int_{\partial \mathcal{P}_i} \left( \frac{\partial \mathbf{q}}{\partial w_i} \cdot n_{\partial \mathcal{P}_i}(\mathbf{q}) \right) \lambda(\mathbf{q}) \, d\mathbf{q},
\end{equation}
where $n_{\partial \mathcal{P}_i}(\mathbf{q})$ represents the unit outward normal to the boundary of the partition, $\partial \mathcal{P}_i$, at point $\mathbf{q}$.
It results in
\begin{align}
    \frac{\partial \lambda_{\mathcal{P}_i}}{\partial w_i} & = \sum_{j \in N_i} \int_{\Delta_{ij}} \left( \frac{\partial \mathbf{q}}{\partial w_i} \cdot n_{ij}(\mathbf{q}) \right) \lambda(\mathbf{q}) \, d\mathbf{q} \nonumber \\
    & \quad  + \sum_{j \in N_i} \int_{\Delta_{i}^{\mathcal{Q}_f}} \left( \frac{\partial \mathbf{q}}{\partial w_i} \cdot n_{ij}(\mathbf{q}) \right) \lambda(\mathbf{q}) \, d\mathbf{q},
    \label{Eq:dlambda_dwi}
\end{align}
where $\Delta_i^{\mathcal{Q}_f}$ is the boundary between the partition and the environment and $n_{ij}(\mathbf{q})$ is the unit normal outward this boundary or the boundary between partitions $i$ and $j$, $\Delta_{ij}$.
The second term is always zero either because $\Delta_i^{\mathcal{Q}_f} = \emptyset$ or because a variation on the weight does not affect the boundary between the partition and the environment, i.e., $\partial \mathbf{q} / \partial w_i = 0$.
Applying the same procedure to $\partial \lambda_{\mathcal{P}_j} / \partial w_i$ and introducing in~\eqref{Eq:dH_dwi_0}, we have
\begin{align}
    \frac{\partial H}{\partial w_i} & = - \frac{1}{\lambda^2_{\mathcal{P}_i}} \sum_{j \in N_i} \int_{\Delta_{ij}} \left( \frac{\partial \mathbf{q}}{\partial w_i} \cdot n_{ij}(\mathbf{q}) \right) \lambda(\mathbf{q}) \, d\mathbf{q} \nonumber \\
    & \quad - \sum_{j \in N_i} \frac{1}{\lambda^2_{\mathcal{P}_j}} \int_{\Delta_{ij}} \left( \frac{\partial \mathbf{q}}{\partial w_i} \cdot n_{ji}(\mathbf{q}) \right) \lambda(q) \, d\mathbf{q}.
\end{align}

Noting that $n_{ji}(\mathbf{q}) = -n_{ij}(\mathbf{q})$, we only have to calculate the scalar product $\partial \mathbf{q} /\partial w_i \cdot n_{ij}(\mathbf{q})$.
The first term can be obtained by deriving the condition of the boundary points~\eqref{Eq:BoundaryPoints} with respect to the power weight,
\begin{equation}
    \frac{\partial}{\partial w_i} \big( d_g(\mathbf{q},\mathbf{g}_i)^2 - w_i \big) = \frac{\partial}{\partial w_i} \big( d_g(\mathbf{q},\mathbf{g}_j)^2 - w_j\big),
\end{equation}
that leads to
\begin{multline}
    2 d_g(\mathbf{q},\mathbf{g}_i) \frac{\partial d_g(\mathbf{q},\mathbf{g}_i)}{\partial \mathbf{q}} \cdot \frac{\partial \mathbf{q}}{\partial w_i} - 1 = \\ 2 d_g(\mathbf{q},\mathbf{g}_j) \frac{\partial d_g(\mathbf{q},\mathbf{g}_j)}{\partial \mathbf{q}} \cdot \frac{\partial \mathbf{q}}{\partial w_i}.
    \label{Eq:df_dwi_1}
\end{multline}
The partial derivative of the geodesic distance with respect to the boundary point $\mathbf{q}$ only affects the first term on the right hand side of~\eqref{Eq:GeodDistDecomposition}:
\begin{equation}
    \frac{\partial d_g(\mathbf{q},\mathbf{g}_i)}{\partial \mathbf{q}} = \frac{\mathbf{q} - h^1_{\mathbf{q},\mathbf{g}_i}}{\| \mathbf{q} - h^1_{\mathbf{q},\mathbf{g}_i} \|}.
    \label{Eq:ddgi_dx}
\end{equation}
Analogously for the other generator we have
\begin{equation}
    \frac{\partial d_g(\mathbf{q},\mathbf{g}_j)}{\partial \mathbf{q}} = \frac{\mathbf{q} - h^1_{\mathbf{q},\mathbf{g}_j}}{\| \mathbf{q} - h^1_{\mathbf{q},\mathbf{g}_j} \|}.
    \label{Eq:ddgj_dx}
\end{equation}
Note that these derivatives have two components, $x$ and $y$, and that they appear in a scalar product in~\eqref{Eq:df_dwi_1}. Therefore, introducing~\eqref{Eq:ddgi_dx} and~\eqref{Eq:ddgj_dx} in~\eqref{Eq:df_dwi_1}, we have
\begin{align}
    & 2 d_g(\mathbf{q},\mathbf{g}_i) \Bigg( \frac{(\mathbf{q} - h^1_{\mathbf{q},\mathbf{g}_i})_x}{\| \mathbf{q} - h^1_{\mathbf{q},\mathbf{g}_i} \|} \frac{\partial \mathbf{q}_x}{\partial w_i} \nonumber \\
    & \hspace{3cm} + \frac{(\mathbf{q} - h^1_{\mathbf{q},\mathbf{g}_i})_y}{\| \mathbf{q} - h^1_{\mathbf{q},\mathbf{g}_i} \|} \frac{\partial \mathbf{q}_y}{\partial w_i} \Bigg) - 1 \nonumber \\
    & = 2 d_g(\mathbf{q},\mathbf{g}_j) \Bigg( \frac{(\mathbf{q} - h^1_{\mathbf{q},\mathbf{g}_j})_x}{\| \mathbf{q} - h^1_{\mathbf{q},\mathbf{g}_j} \|} \frac{\partial \mathbf{q}_x}{\partial w_i} \nonumber \\
    & \hspace{3cm} + \frac{(\mathbf{q} - h^1_{\mathbf{q},\mathbf{g}_j})_y}{\| \mathbf{q} - h^1_{\mathbf{q},\mathbf{g}_j} \|} \frac{\partial \mathbf{q}_y}{\partial w_i} \Bigg)
\end{align}
and, regrouping the terms,
\begin{align}
    \Bigg[ & 2 d_g(\mathbf{q},\mathbf{g}_i) \frac{(\mathbf{q} - h^1_{\mathbf{q},\mathbf{g}_i})_x}{\| \mathbf{q} - h^1_{\mathbf{q},\mathbf{g}_i} \|} - 2 d_g(\mathbf{q},\mathbf{g}_j) \frac{(\mathbf{q} - h^1_{\mathbf{q},\mathbf{g}_j})_x}{\| \mathbf{q} - h^1_{\mathbf{q},\mathbf{g}_j} \|}, \nonumber \\ 
    & 2 d_g(\mathbf{q},\mathbf{g}_i) \frac{(\mathbf{q} - h^1_{\mathbf{q},\mathbf{g}_i})_y}{\| \mathbf{q} - h^1_{\mathbf{q},\mathbf{g}_i} \|} \nonumber \\
    & \hspace{1.5cm} - 2 d_g(\mathbf{q},\mathbf{g}_j) \frac{(\mathbf{q} - h^1_{\mathbf{q},\mathbf{g}_j})_y}{\| \mathbf{q} - h^1_{\mathbf{q},\mathbf{g}_j} \|} \Bigg] \cdot \frac{\partial \mathbf{q}}{\partial w_i}= 1.
    \label{Eq:df_dwi_2}
\end{align}

To obtain the normal to the boundary between two partitions, $n_{ij}(\mathbf{q})$, we make use of the property of the gradient of a function, that is always perpendicular to the level curves of the function.
To this end, we transform the equation that defines the boundary points~\eqref{Eq:BoundaryPoints} into a function,
\begin{equation}
    f(\mathbf{q}) = d_g(\mathbf{q},\mathbf{g}_i)^2 - w_i - d_g(\mathbf{q},\mathbf{g}_j)^2 + w_j.
\end{equation}
Then, we calculate the gradient at the points that satisfy $f(\mathbf{q}) = 0$, that is, $\mathbf{q} \in \Delta_{ij}$, for the $x$ component,
\begin{align}
    & \frac{\partial f(\mathbf{q})}{\partial \mathbf{q}_x} =
    2 d_g(\mathbf{q},\mathbf{g}_i) \frac{\partial d_g(\mathbf{q},\mathbf{g}_i)}{\partial \mathbf{q}_x} - 2 d_g(\mathbf{q},\mathbf{g}_j) \frac{\partial d_g(\mathbf{q},\mathbf{g}_j)}{\partial \mathbf{q}_x} \nonumber \\
    & \quad = 2 d_g(\mathbf{q},\mathbf{g}_i) \frac{\partial \| \mathbf{q} - h^1_{\mathbf{q},\mathbf{g}_i} \|}{\partial \mathbf{q}_x} - 2 d_g(\mathbf{q},\mathbf{g}_j) \frac{\partial \| \mathbf{q} - h^1_{\mathbf{q},\mathbf{g}_j} \|}{\partial \mathbf{q}_x}  \nonumber \\
    & \quad = 2 d_g(\mathbf{q},\mathbf{g}_i) \frac{(\mathbf{q} - h^1_{\mathbf{q},\mathbf{g}_i})_x}{\| \mathbf{q} - h^1_{\mathbf{q},\mathbf{g}_i} \|} - 2 d_g(\mathbf{q},\mathbf{g}_j) \frac{(\mathbf{q} - h^1_{\mathbf{q},\mathbf{g}_j})_x}{\| \mathbf{q} - h^1_{\mathbf{q},\mathbf{g}_j} \|},
    \label{Eq:df_dxk}
\end{align}
and, similarly, for the $y$ component.
Therefore, the normal to the boundary is
\begin{equation}
    n_{ij}'(\mathbf{q}) = \left(\frac{\partial f(\mathbf{q})}{\partial \mathbf{q}_x},\frac{\partial f(\mathbf{q})}{\partial \mathbf{q}_y} \right)
\end{equation}
and the unit normal is
\begin{equation}
    n_{ij}(\mathbf{q}) = \frac{n_{ij}'(\mathbf{q})}{\|n_{ij}'(\mathbf{q})\|}.
\end{equation}
Introducing the last three equations in~\eqref{Eq:df_dwi_2}, we obtain
\begin{equation}
\label{Eq:invn}
    \frac{\partial \mathbf{q}}{\partial w_i} \cdot n_{ij}(\mathbf{q}) = \dfrac{1}{\|n_{ij}'(\mathbf{q})\|},
\end{equation}
and replacing in~\eqref{Eq:dlambda_dwi} we obtain the complete formulation of the gradient stated in~\eqref{Eq:dH_dwi}.
Finally, the proof of the convergence is equivalent to Theorem 3.7 in~\cite{pavone2011distequitpartition}, completing the proof of this theorem.
\end{proof}

\begin{remark}
    The main differences of our proof with respect to the one presented in~\cite{pavone2011distequitpartition} reside in the derivation of the geodesic distance and in the calculation of the normal to the boundary points. These two key modifications generalize the algorithm to any kind of non-convex environments, yet still giving the same solution for convex ones.
\end{remark}

It is also important to highlight that the gradient is still distributed since each robot can update its own weight and, therefore, its own partition, by only communicating with its neighbors. Additionally, note that the gradient has no physical representation in the real environment since it only affects the weights associated with the generators.

\section{Partition Enhancements}
\label{sec:connectivity}
The partitions obtained with the algorithm described in the previous section are proven to be equitable. However, for non-convex environments, especially for complex ones, they might be disconnected, in the sense that 
in order to reach some points of the partition, a robot needs to go through the partitions of other robots.
This may affect significantly the persistent coverage of the environment.
Moreover, they do not take into account the different capabilities that each robot might have.
In this section we present two enhancements of our partitioning algorithm: one where we control the positions of the generators to reduce possible disconnections and other to account for energy and coverage power of the robots.

\subsection{Connectivity-Aware Partitioning}

Let us begin by defining the connected components, $\mathcal{P}_i^\ell(\mathbf{w})$, $\ell \in \{1, \dots, L_i\}$, $L_i\in \mathbb{Z}_+$, that form a partition:
\begin{equation}
    \mathcal{P}_i = \bigcup_{\ell = 1}^{L_i}  \mathcal{P}_i^\ell.
\end{equation}
Between these connected components we select the one with the highest workload:
\begin{equation}
\mathcal{P}_i^c = \argmax_{\mathcal{P}_i^\ell} \lambda_{\mathcal{P}_i^\ell}.
\end{equation}
Note that for connected partitions $\mathcal{P}_i^c$ coincides with $\mathcal{P}_i$.
Then, we calculate the center of mass of $\mathcal{P}_i^c$ as
\begin{equation}
    \mathbf{g}_i^* = \frac{1}{\lambda_{\mathcal{P}_i^c}} \int_{\mathcal{P}_i^c} \mathbf{q} \, \lambda(\mathbf{q}) \, d\mathbf{q}.
\end{equation}
This point is the best point of the environment in which the generator $\mathbf{g}_i$ could be located to favor the connectivity of its partition. However, it cannot be moved straightaway in the direction $\overline{\mathbf{g}_i\mathbf{g}_i^*}$ since there are two restrictions that must be taken into account. The first one is that the motion of the generator must not hinder the evolution of the weights towards the equitable partition. Therefore, such motion can only be executed if it contributes to the minimization of~\eqref{Eq:CostfunctionPartition} or, at least, it is not detrimental.
The second restriction is that in a non-convex environment it has to follow the geodesic path from its current position to the center of mass to avoid sudden changes in the shape, size and workload of the partitions. If we call $\gamma_i$ the shortest path from $\mathbf{g}_i$ to $\mathbf{g}_i^*$, we want the generator to move in the direction $\gamma_i(\mathbf{g}_i)$, that is the direction of the path at $\mathbf{g}_i$.
In this context, the algorithm that we propose is presented in the following theorem.
\begin{theorem}
    The power diagram generated by $\mathcal{G}$ and $\mathbf{w}$ converges to an equitable power diagram 
    under the distributed control law
    \begin{subequations}
        \begin{align}
            \dot{w}_i & = - k_w \, \frac{\partial H}{\partial w_i}, \label{Subeq:ControlLawW}\\
            \dot{\mathbf{g}}_i & = k_g \, f_{sat}\left( \overline{\mathbf{g}_i\mathbf{g}_i^*} \cdot \frac{-\partial H}{\partial \mathbf{g}_i} \right) \, \gamma_i(\mathbf{g}_i), \label{Subeq:ControlLawG}
        \end{align}
        \label{Eq:ControlLaw2}
    \end{subequations}
    with $k_w, k_g$, positive gains,
    \begin{equation}
        f_{sat}(x) = \left\{ 
          \begin{aligned}
            & 0, &\forall \, x \leq 0, \\
            \exp\bigg( & \frac{-1}{(k_{sat} \, x )^2} \bigg), & \forall \, x > 0,
          \end{aligned}
          \right.
          \label{Eq:Saturation}
    \end{equation}
    a saturation function with $k_{sat} > 0$, and
    \begin{multline}
        \frac{\partial H}{\partial \mathbf{g}_{i,k}} = \sum_{j \in N_i} \left( \frac{1}{\lambda^2_{\mathcal{P}_j}} - \frac{1}{\lambda^2_{\mathcal{P}_i}} \right) \\
         \int_{\Delta_{ij}}  \frac{(\mathbf{h}_{\mathbf{q},\mathbf{g}_i}^{l_{\mathbf{q},\mathbf{g}_i}} - \mathbf{g}_i)_k}{\| \mathbf{h}_{\mathbf{q},\mathbf{g}_i}^{l_{\mathbf{q},\mathbf{g}_i}} - \mathbf{g}_i \| } \, \dfrac{\lambda(\mathbf{q})}{\|n_{ij}'(\mathbf{q})\|} \, d\mathbf{q},
         \label{Eq:dH_dgik}
    \end{multline}
    where $n_{ij}'(\mathbf{q})$ is the outward normal to the boundary between the partitions $i$ and $j$ at point $\mathbf{q}$.
\label{Theo:EquitConn}
\end{theorem}
\begin{proof}
The convergence to an equitable power diagram is guaranteed under the first part of the control law,~\eqref{Subeq:ControlLawW}, in Theorem~\ref{Theo:Equitable}. In this proof we only show that the second part of the control law,~\eqref{Subeq:ControlLawG}, does not counteract this convergence.

The direction of motion of the generators that maximizes the improvement of the cost function is the gradient of the cost function with respect to $\mathbf{g}_i$, i.e., $\partial H / \partial \mathbf{g}_i$.
This gradient is obtained in the same way as with respect to $w_i$ in the proof of Theorem~\ref{Theo:Equitable} with only two particularities.
The derivatives of the geodesic distances with respect to $\mathbf{g}_{i,x}$ are
\begin{align}
    \frac{\partial d_g(\mathbf{q},\mathbf{g}_i)}{\partial \mathbf{g}_{i,x}} & = \frac{\partial \| \mathbf{q} - \mathbf{h}_{\mathbf{q},\mathbf{g}_i}^1 \|}{\partial \mathbf{g}_{i,x}} + \dots + \frac{\partial \| \mathbf{h}_{\mathbf{q},\mathbf{g}}^{l_{\mathbf{q},\mathbf{g}_i}} - \mathbf{g}_i \|}{\partial \mathbf{g}_{i,x}} \nonumber \\
    & = \frac{\mathbf{q} - \mathbf{h}_{\mathbf{q},\mathbf{g}_i}^1}{\| \mathbf{q} - \mathbf{h}_{\mathbf{q},\mathbf{g}_i}^1 \|}  \cdot \frac{\partial \mathbf{q}}{\partial \mathbf{g}_{i,x}} - \frac{ (\mathbf{h}_{\mathbf{q},\mathbf{g}}^{l_{\mathbf{q},\mathbf{g}_i}} - \mathbf{g}_i)_x}{\| \mathbf{h}_{\mathbf{q},\mathbf{g}}^{l_{\mathbf{q},\mathbf{g}_i}} - \mathbf{g}_i \|},
\end{align}
\begin{gather}
    \frac{\partial d_g(\mathbf{q},\mathbf{g}_j)}{\partial \mathbf{g}_{i,x}} = \frac{\partial \| \mathbf{q} - \mathbf{h}_{\mathbf{q},\mathbf{g}_j}^1 \|}{\partial \mathbf{g}_{i,x}} = \frac{\mathbf{q} - \mathbf{h}_{\mathbf{q},\mathbf{g}_j}^1}{\| \mathbf{q} - \mathbf{h}_{\mathbf{q},\mathbf{g}_j}^1 \|} \cdot \frac{\partial \mathbf{q}}{\partial \mathbf{g}_{i,x}},
\end{gather}
and equivalent for $\mathbf{g}_{i,y}$.
Note that in the first step of the first derivative all the intermediate terms represented with the dots are equal to zero since they do not depend on $\mathbf{g}_{i,x}$ or $\mathbf{q}$.
The second particularity is that in the end we obtain
\begin{equation}
    \frac{\partial \mathbf{q}}{\partial  \mathbf{g}_{i,x}} \cdot n_{ij} = \dfrac{1}{\|n_{ij}'(\mathbf{q})\|} \frac{(\mathbf{h}_{\mathbf{q},\mathbf{g}}^{l_{\mathbf{q},\mathbf{g}_i}} - \mathbf{g}_i)_x}{\| \mathbf{h}_{\mathbf{q},\mathbf{g}}^{l_{\mathbf{q},\mathbf{g}_i}} - \mathbf{g}_i \|},
\end{equation}
which leads to~\eqref{Eq:dH_dgik}.

The saturation function in~\eqref{Subeq:ControlLawG} compares the direction of the gradient with the direction from the generator to the centroid of the biggest connected component. If both directions are approximately aligned, i.e., the movement of the generator to the centroid favors the convergence to the equitable partition, the generator is able to follow the geodesic path to the centroid according to $\gamma_i(\mathbf{g}_i)$. Otherwise, if the movement to the centroid hinders the convergence, $f_{sat} = 0$ and the generator does not move.
\qed
\end{proof}

\begin{remark}
Even though Theorem~\ref{Theo:EquitConn} does not provide formal guarantees on connectedness, since the movement of the generator is towards the centroid of the component with the biggest workload, we can assert that this partition methodology favors connectivity of the regions.
\end{remark}

Although the generators move towards their respective centroids, the convergence of the control law is to an equitable partition. Therefore, the generator may either reach the centroids or stop somewhere in the geodesic path to such point when an equitable partition is achieved.

It should also be noted that the motion of the generators in this control law is independent on the actual motion of the robots, which will be discussed in the Section~\ref{sec:coverage}. This also means that, if the computed partition is not satisfactory, e.g., is still disconnected, the algorithm can be run again with different initial conditions until a connected partition is obtained.
Similarly, the robots can ``trade'' small regions of equivalent workload so that equity is preserved, but in such a way that the partitions are improved, e.g., connected or associated to physical rooms.
We provide a more detailed empirical analysis in Section~\ref{sec:simulations} where we analyze how often the resulting partitions are not connected, leaving more sophisticated partition mechanisms, like trading, for future work.

\subsection{Capability-Aware Partitioning}

The algorithm to find the equitable partition that we have introduced simply divides the environment in regions with the same workload. 
It is noteworthy that the cost function can be easily extended with gains that reflect different coverage capabilities for each robot.
If we consider
\begin{equation}
    H(\mathbf{w}) = \sum_{i=1}^N \frac{c_i}{\lambda_{\mathcal{P}_i(\mathbf{w})}},
\end{equation}
as an alternative cost function, where $c_i$ is a constant value.
Since this term is constant with respect to $w_i$, we can still use the control law of the previous section. The only difference is that each robot now will have a different gain driving the evolution of its own partition weight and generator.
The partitions obtained in this case will satisfy that
\begin{equation}
    \dfrac{c_i}{\lambda_{\mathcal{P}_i(\mathbf{w})}} = \dfrac{c_j}{\lambda_{\mathcal{P}_j(\mathbf{w})}}.
\end{equation}
As an example, the constants, $c_i$ can be chosen by
\begin{equation}
    c_i = \displaystyle\int_{\mathbf{\Omega}_i(\mathbf{p})} \alpha_i(\mathbf{q},\mathbf{p}) \, \mathrm{d}\mathbf{q},
\end{equation}
where $\mathbf{p}$ is a virtual position to integrate the maximum coverage that each agent can provide.
With this value, the partition of a robot that is able to produce more coverage action will be bigger than the partition of another robot that is able to produce less action.

\section{Finite-Horizon, Graph-based Coverage Planning}
\label{sec:coverage}

The second part of our persistent solution is in charge of planning online coverage paths for the robots.
To do this, we first define the coverage error at each location and time instant,
\begin{equation}
    e(\mathbf{q},k) = \Phi(\mathbf{q}) \, \big(Z^*(\mathbf{q}) - Z(\mathbf{q},k) \big),
    \label{Eq:CovError}
\end{equation}
where we allow positive values for under-covered points and negative values for over-covered ones.
We assume that the initial coverage value for each point is equal to zero.
Similarly, we define the \emph{quadratic} coverage error over the whole environment as
\begin{equation}
f(k) = \int_{\mathcal{Q}} e^2(\mathbf{q},k) d\mathbf{q},
\end{equation}
in this case to emphasize that it is equally bad to be above and below the desired coverage level.

Ideally, the best coverage paths would be the $\Gamma_i$ that minimize $f(k)$ for the whole duration of the coverage task, e.g., minimize $\sum_k f(k)$. However, due to the coverage deterioration over time, finding a solution to this optimization problem is computationally intractable even for small teams, as shown in~\cite{palacios2016branchandbound}.
Therefore, in the following we are going to propose an online planning solution that minimizes the quadratic coverage error of a subset of points in $f(k)$ at each planning time.

In particular, we are going to generate a graph of locations that considers the coverage capabilities of each robot inside the previously computed equitable partitions.
This way, we can reduce the planning problem to that of finding the shortest path in a graph, which can be computed in polynomial time very efficiently.
Besides, the construction of the graph favors the planning of sweep-like paths, that are very suitable for coverage problems.
In the rest of the section we explain in detail the whole procedure.


\subsection{Graph Construction based on Sweep-like Paths}
\label{Subsec:Graph}

Let us first define $\mathcal{P}_i'$ as the subset of locations from $\mathcal{P}_i$ where the robot can be located without colliding with an obstacle.

Inside the partitions we separate the construction of the path graph in two parts: grid and boundary.
The aim of the first part is to define paths that allow the robot to perform a complete sweep of the partition and the second part is intended to cover the entire boundary of the partition.

\begin{figure}[!ht]
        \centering
        \subfloat[Grid.]{ \includegraphics[width=0.23\textwidth]{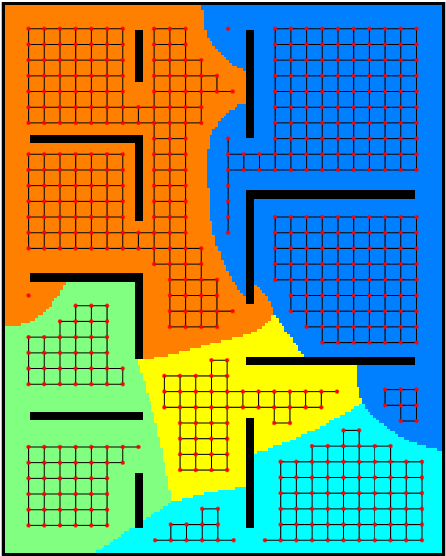}
                \label{Subfig:GridNodes}
                }
                \hfil
        \subfloat[Boundary.]{ \includegraphics[width=0.23\textwidth]{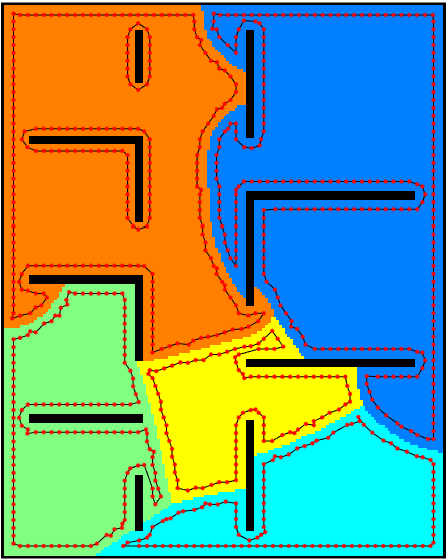}
                \label{Subfig:BoundNodes}
                }
                
        \subfloat[Complete graph.]{ \includegraphics[width=0.23\textwidth]{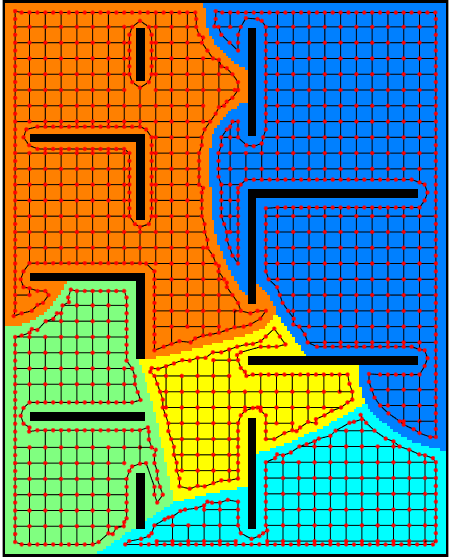}
                \label{Subfig:Graph}
                }
    \label{Fig:GraphConstruction}
    \caption{Example of graph construction.}
\end{figure}

For the first part, we intersect a grid of points with the partition: 
\begin{equation}
	\mathcal{V}^g_i = \{ \mathbf{v} \in \mathcal{P}_i' \, | \, \mathbf{v} = ( x_i^0 + k \, d_i, y_i^0 \pm \ell \, d_i ), k, \ell \in \mathbb{N}_0 \},
\end{equation}
where $\mathbf{v}_i^0 = ( x_i^0, y_i^0 ) \in \mathcal{P}_i'$ is the
left-most point of $\mathcal{P}_i'$ and $d_i$ is the optimal separation between consecutive sweeping lines.
Note that the selection of $\mathbf{v}_i^0$ is arbitrary since the grid is fixed by $d_i$ and the boundary is covered by the second part of the graph.
The vertices in $\mathcal{V}^g_i$ are connected by an edge if they are at a distance equal to $d_i$.
Formally,
\begin{equation}
	\mathcal{E}^g_i = \{ (\mathbf{v}_1, \mathbf{v}_2) \, | \, \| \mathbf{v}_1 - \mathbf{v}_2 \| = d_i, \, \forall \, \mathbf{v}_1, \mathbf{v}_2 \in \mathcal{V}^g_i \}.
\end{equation}
Figure~\ref{Subfig:GridNodes} shows an example of the vertices and edges that correspond to the grid part of the path graph.

The optimal separation between two horizontal or two vertical sweeping lines, $d_i$, depends on the shape of the coverage area and the production function.
It can be obtained by minimizing
\begin{multline}
	\int_{-r_i^{cov}}^{r_i^{cov}} \int_{-r_i^{cov}}^{d_i+r_i^{cov}} \Big( \overline{\alpha} - \big(\alpha_i((x,y),(0,0)) \\ 
	+ \alpha_i((x,y),(0,d_i)) \big) \Big)^2 dy \, dx,
	\label{Eq:Optimdi}
\end{multline}
with
\begin{equation}
    \overline{\alpha} = 2 \, \frac{\int_{\Omega_i(\mathbf{p}_i)} \alpha_i(\mathbf{p}_i) d\mathbf{q}}{\int_{-r_i^{cov}}^{r_i^{cov}} \int_{-r_i^{cov}}^{d_i+r_i^{cov}} dy \, dx}.
\end{equation}

\begin{figure}[!ht]
    \centering
    \includegraphics[width=0.35\textwidth]{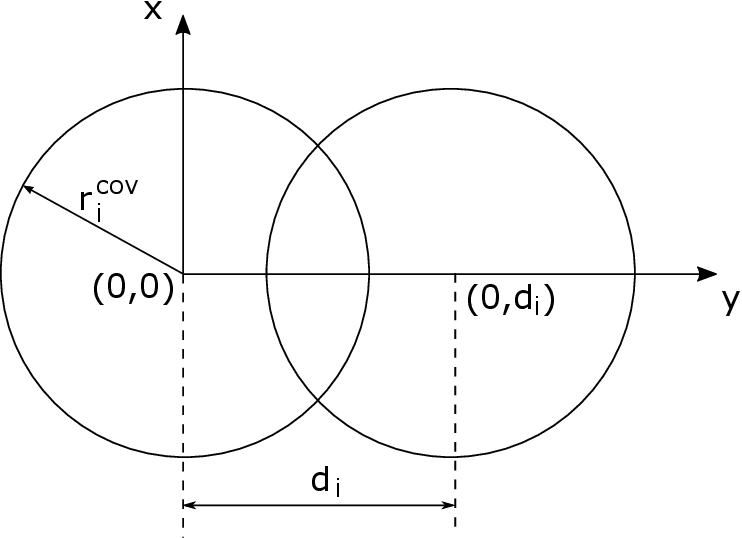}
    \caption{Schematic of the overlap of the coverage areas for two consecutive sweeps in the $x$ direction and the optimal separation between them.}
    \label{Fig:d_i}
\end{figure}

The idea behind~\eqref{Eq:Optimdi} is illustrated in Fig.~\ref{Fig:d_i}. The two circumferences represent the areas that the robot can cover when it is located at $\mathbf{p}_i^1 = (0,0)$ and $\mathbf{p}_i^2 = (0,d_i)$, respectively. 
The two locations correspond to consecutive sweeps in the $x$ direction, i.e. two vertices in $\mathcal{V}^g_i$ connected by an edge in $\mathcal{E}^g_i$. 
Note that $\Omega_i(\mathbf{p}_i)$ does not have to be circular, but only bounded by the circumference, and that $\alpha_i(\mathbf{p}_i)$ can be any function. 
The optimal distance between these parallel sweeps, $d_i$, is the one that optimizes the sum of the coverage provided from these two positions. In particular, the goal of~\eqref{Eq:Optimdi} is that the same coverage is provided to each point that can be covered by the robot from any of the positions. This coverage is represented by $\overline{\alpha}$, that is the sum of the coverage from the two positions divided by the total area. Therefore, $\overline{\alpha}$ can also be seen as an average coverage level. Eq.~\eqref{Eq:Optimdi} computes the quadratic difference between the coverage provided to each point, i.e., the sum of $\alpha_i$ from $p_i^1$ and $p_i^2$, and this average to obtain a $d_i$ that keeps the coverage of all points as close as possible to $\overline{\alpha}$.

For the second part of the graph, we represent the boundary as $\partial\mathcal{P}_i'$ and locate a finite number of vertices over it as follows:
\begin{multline}
	\mathcal{V}^b_i = \{ \mathbf{v} \in \partial\mathcal{P}_i' \, | \\
	\, \mathbf{v}_x = x_i^0 \pm k \, \frac{d_i}{2} \lor \mathbf{v}_y = y_i^0 \pm \ell \, \frac{d_i}{2}, k, \ell \in \mathbb{Z} \}.
\end{multline}
They are located at the intersections of a square grid of size $d_i / 2$ with the boundary, to provide a sufficiently fine discretization.
These vertices are linked by the edges
\begin{equation}
	\mathcal{E}^b_i = \{ (\mathbf{v}_1, \mathbf{v}_2) \, | \,  \mathbf{v}_1 \text{ and } \mathbf{v}_2 \text{ are consecutive over } \partial\mathcal{P}_i' \}.
\end{equation}
These edges allow the robots to follow the entire boundary of the partition as shown in Fig.~\ref{Subfig:BoundNodes}.

Finally, we merge the two parts by including an additional set of links,
\begin{align}
	\mathcal{E}^{gb}_i = \{ & (\mathbf{v}_1, \mathbf{v}_2) \, | \,  |\mathbf{v}_1^x - \mathbf{v}_2^x| < d_i \, \land \, |\mathbf{v}_1^y - \mathbf{v}_2^y| = 0 \, \lor \nonumber \\
	& |\mathbf{v}_1^x - \mathbf{v}_2^x| = 0 \, \land \, |\mathbf{v}_1^y - \mathbf{v}_2^y| < d_i, \nonumber \\
	& \qquad \forall \, \mathbf{v}_1 \in \mathcal{V}^g_i, \mathbf{v}_2 \in \mathcal{V}^b_i \}.
\end{align}
As can be seen in Fig.~\ref{Subfig:Graph}, these edges link the vertices of both parts that are closer than $d_i$ either in the $x$- or in the $y$-axis and the resulting directed path graph can be expressed as
\begin{equation}
	\mathcal{G}_i = ( \mathcal{V}_i, \mathcal{E}_i)  \equiv ( \mathcal{V}^g_i \cup \mathcal{V}^b_i, \mathcal{E}^g_i \cup \mathcal{E}^b_i \cup \mathcal{E}^{gb}_i ).
\end{equation}

Besides the computational savings of considering a discrete subset of positions to compute the coverage paths, this simplification can also be exploited to decompose the general planning problem for $N$ robots into $N$ individual problems for one robot each.
\begin{prop}
\label{Prop:division}
The paths that optimize $f(k),$ subject to the constraint of each robot only following trajectories within $\mathcal{G}_i,$
are the paths that optimize individually the cost function inside each partition, i.e.,
\begin{equation}
\label{Eq:PathSeparated}
    \min_{\Gamma_i} f(k) = \min_{\Gamma_1} f_1(k) + \, \dots \, + \min_{\Gamma_N} f_N(k),
\end{equation}
with
\begin{equation}
    f_i(k) = \int_{\mathcal{P}_i} e^2(\mathbf{q},k)\ d\mathbf{q}.
    \label{Eq:PartCostFunc}
\end{equation}
\end{prop}
\begin{proof}
It is straightforward to see that
\begin{equation}
    f(k) = f_1(k) + \, \dots \, + f_N(k).
\end{equation}
Since the path of each robot is constrained to be within the graph $\mathcal{G}_i$, the coverage contribution of such path is also restricted to $\mathcal{P}_i$, because no coverage is applied outside of it.
Then, the path of each robot $i$ only affects to the value of $f_i(k)$ in~\eqref{Eq:PartCostFunc}, which demonstrates~\eqref{Eq:PathSeparated}.
\qed
\end{proof}
The immediate consequence of this proposition is that each robot can plan locally its own path, inside its own partition according to $f_i(k)$, working distributively towards a global minimization of $f(k)$.
This also implies that at this stage robots do not need to share any coverage value with any other robot, since they act independently in their regions.
It should also be noted that Proposition~\ref{Prop:division} is valid for any partition of the environment, as long as the graphs $\mathcal{G}_i$ ensure that each robot can only apply coverage within its assigned partition.
Likewise, the proposition can be applied to other cost functions that, similarly to $f(k)$, consider additive cost in the points of the environment. We are going to utilize this in the next sub-section to obtain online trajectories with a computationally light algorithm.


\subsection{Optimal Path Planning}

In order to compute the trajectories within the graph, we first transform it into a weighted digraph. We assign to each edge a weight equivalent to the amount of coverage required to reach the desired value, $Z^*,$ in the head of the edge,
\begin{equation}
	\omega_i(\mathbf{v}_1,\mathbf{v}_2,k) = \max\big(e(\mathbf{v}_2,k),0\big), 
\end{equation}
for all $(\mathbf{v}_1, \mathbf{v}_2) \in \mathcal{E}_i.$
In the second place, for a given trajectory, $\Gamma_i(k)$, through the graph $\mathcal{G}_i$ at time $k$, starting from the position of robot $i$ at that time, we define the 
error per number of visited vertices as
\begin{equation}
	g_i\big(\Gamma_i(k) \big) = \frac{\sum_{\ell=1}^{L_v} \omega_i(\mathbf{v}_{\ell-1},\mathbf{v}_\ell,k)}{L_v},
\end{equation}
where $L_v$ is the number of traversed vertices in $\Gamma_i$
and $(\mathbf{v}_{\ell-1}, \mathbf{v}_{\ell}) \in \mathcal{E}_i$ for all $\ell$.
Therefore, the optimization problem considered by our planning algorithm is
\begin{equation}
\label{Eq:CostPlanner}
	\underset{\Gamma_i(k)}{\text{maximize}} \quad g_i\big(\Gamma_i(k) \big).
\end{equation}
This optimization problem aims at finding the optimal between all the paths of all possible lengths that go through the graph without repeating vertices.

The transformation of the coverage path planning into finding the optimal path through a graph allows the problem to be solved rapidly and efficiently with proven methods to find the shortest paths in graphs. In particular, we adapt the Bellman-Ford algorithm~\cite{cormen2009algorithms} in two ways.
Instead of the shortest distance, we look for the highest accumulated error per vertex and, therefore, we initialize our metric to $-\infty$ for all the nodes of the graph.
Additionally, since the aim is to maximize this metric, the path to a vertex is updated and the vertex appended to the priority queue if the previously stored value of the metric is lower than the new one.
It should be noted that these modifications do not alter the optimality of the method and maintain the worst-case complexity of $\mathcal{O}(|\mathcal{V}_i| \, |\mathcal{E}_i|)$.
Each path is then fully executed by the robot until it is finished.
Then, the planning process is repeated with the new values of the coverage errors in each vertex.

\section{Simulation Results}
\label{sec:simulations}

\begin{figure}[!ht]
    \centering
    \includegraphics[width=0.3\textwidth]{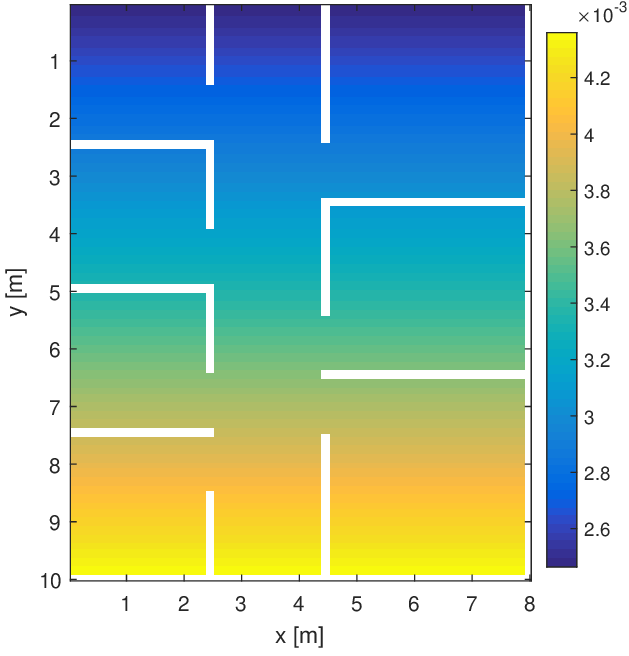}
    \caption{Normalized work function $\lambda(\mathbf{q})$ over the \emph{Open Rooms} environment. White areas represent obstacles.}
    \label{Fig:Workmap}
\end{figure}

\begin{figure*}[!ht]
        \centering
        \subfloat[Open Rooms.]{ \includegraphics[width=0.23\linewidth]{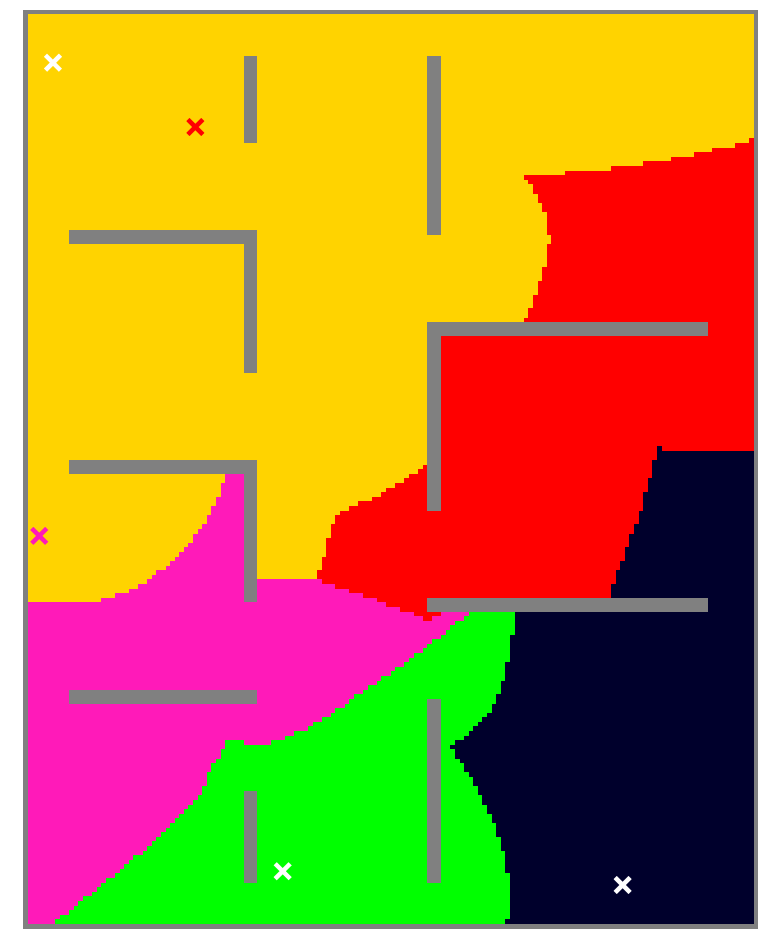}
                \label{Subfig:NoMoveOpenRooms}
                }
                \hfil
        \subfloat[Rooms.]{ \includegraphics[width=0.23\linewidth]{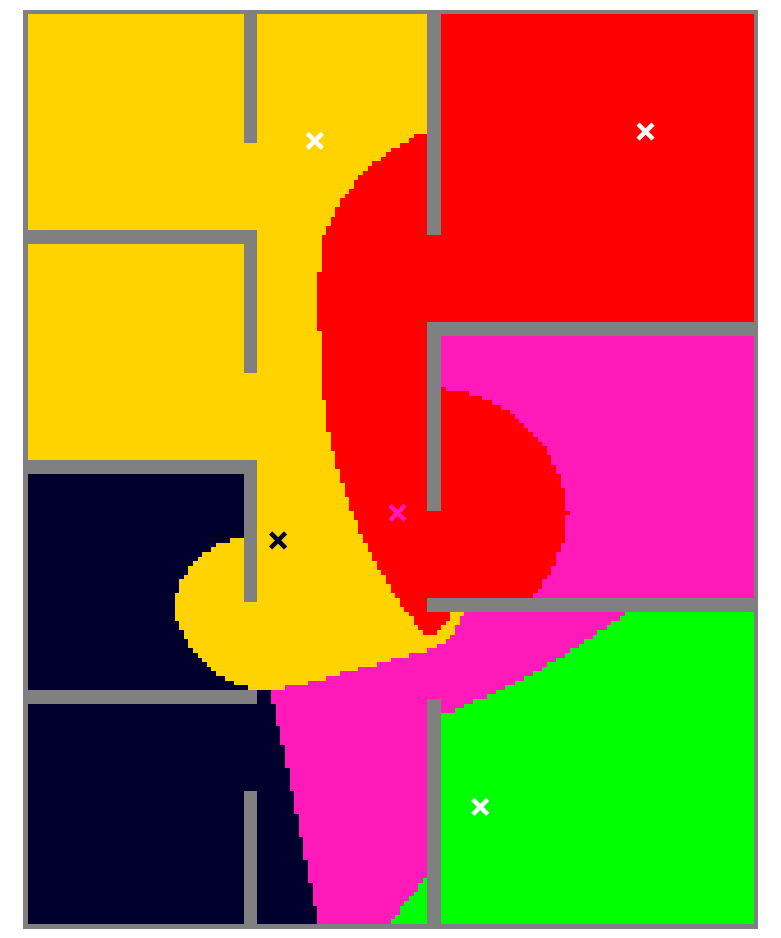}
                \label{Subfig:NoMoveRooms}
                }
                \hfil
        \subfloat[Spiral.]{ \includegraphics[width=0.23\linewidth]{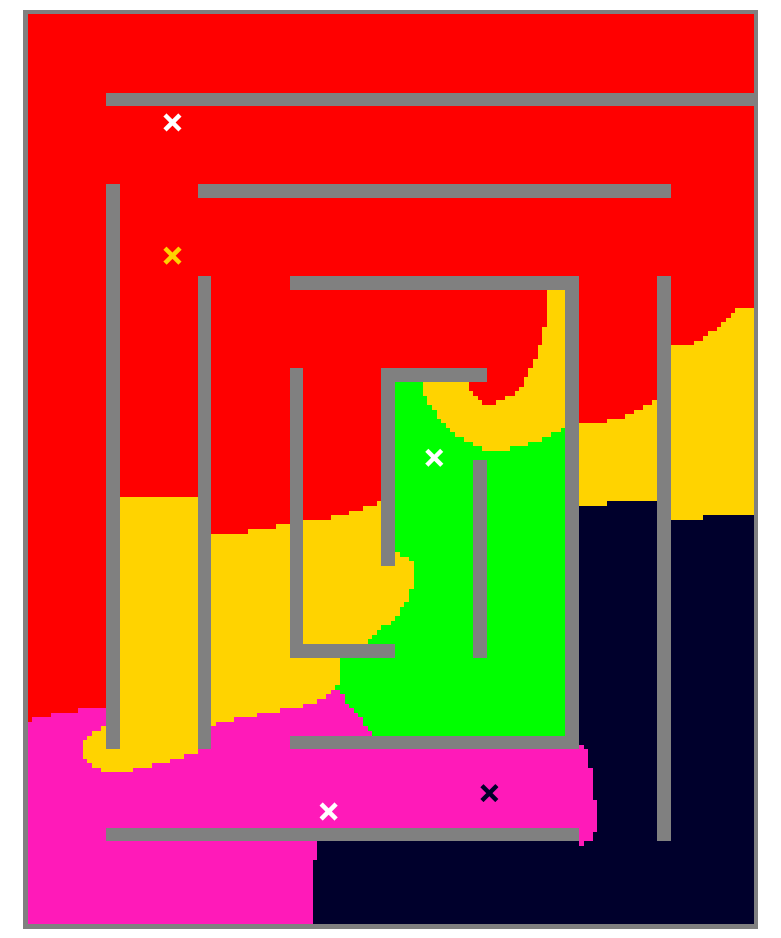}
                \label{Subfig:NoMoveSpiral}
                }
                \hfil
        \subfloat[Maze.]{ \includegraphics[width=0.23\linewidth]{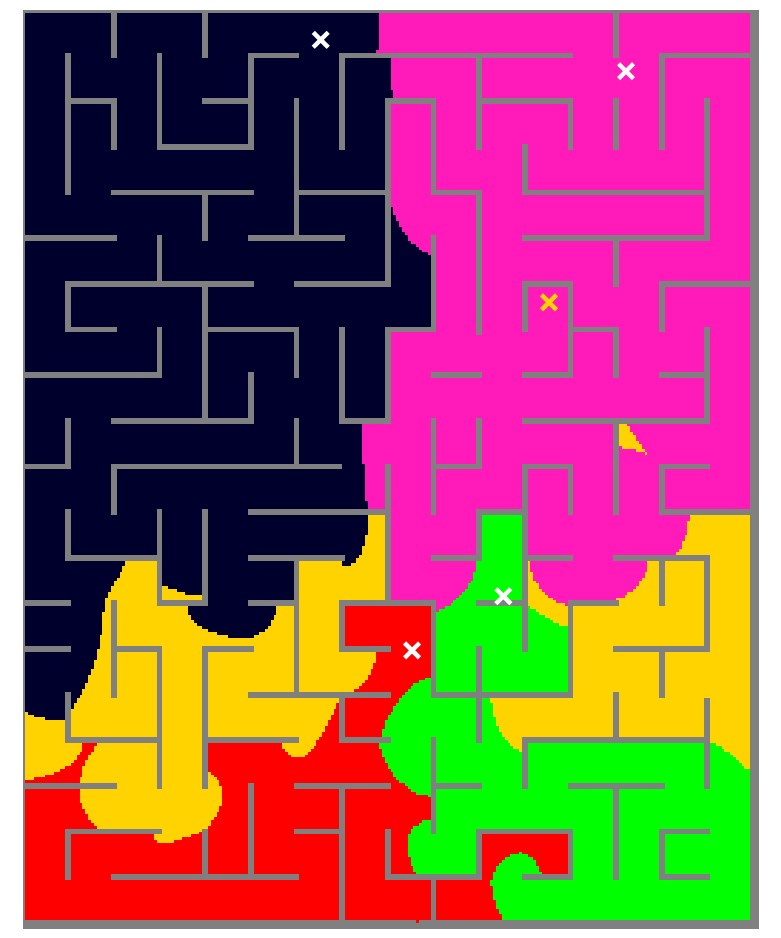}
                \label{Subfig:NoMoveMaze}
                }

        \subfloat[Open Rooms.]{ \includegraphics[width=0.23\linewidth]{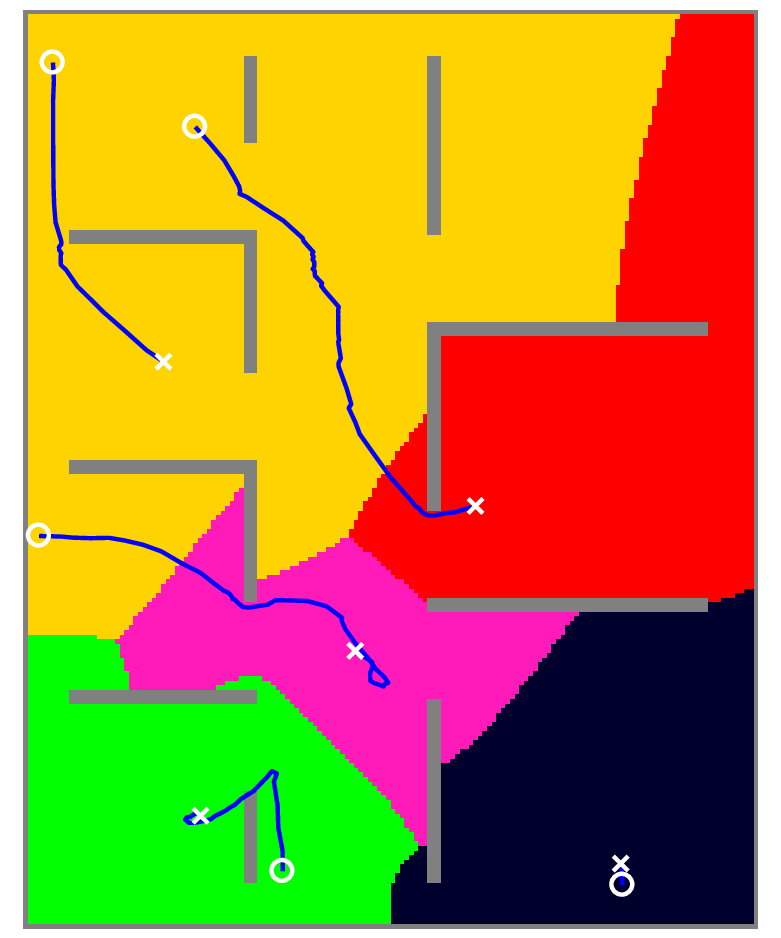}
                \label{Subfig:MoveOpenRooms}
                }
                \hfil
        \subfloat[Rooms.]{ \includegraphics[width=0.23\linewidth]{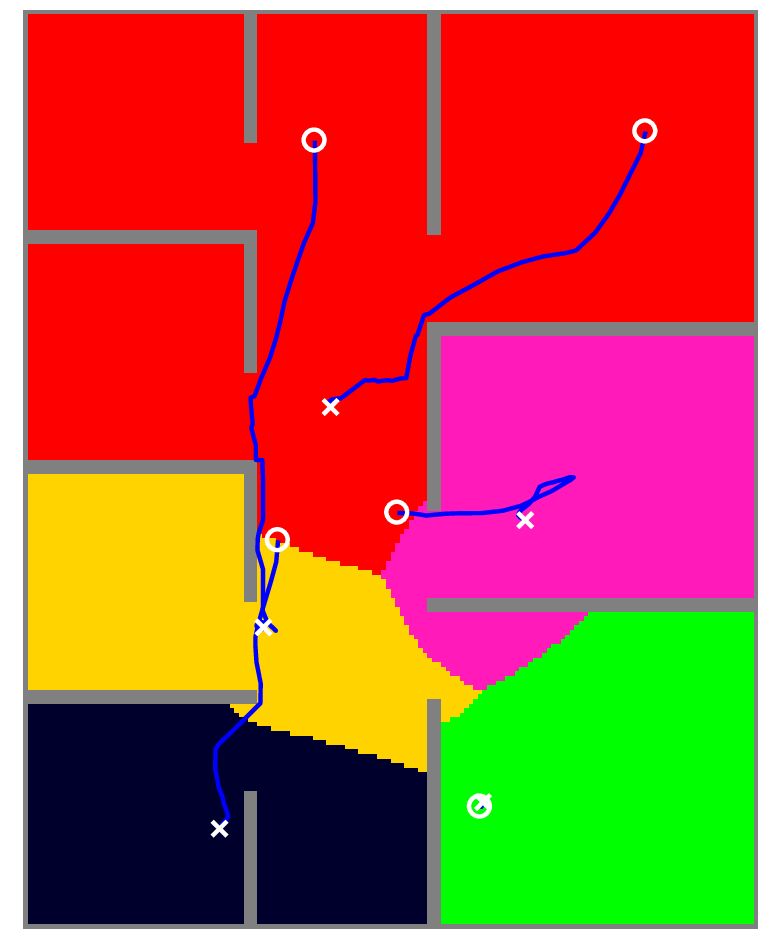}
                \label{Subfig:MoveRooms}
                }
                \hfil
        \subfloat[Spiral.]{ \includegraphics[width=0.23\linewidth]{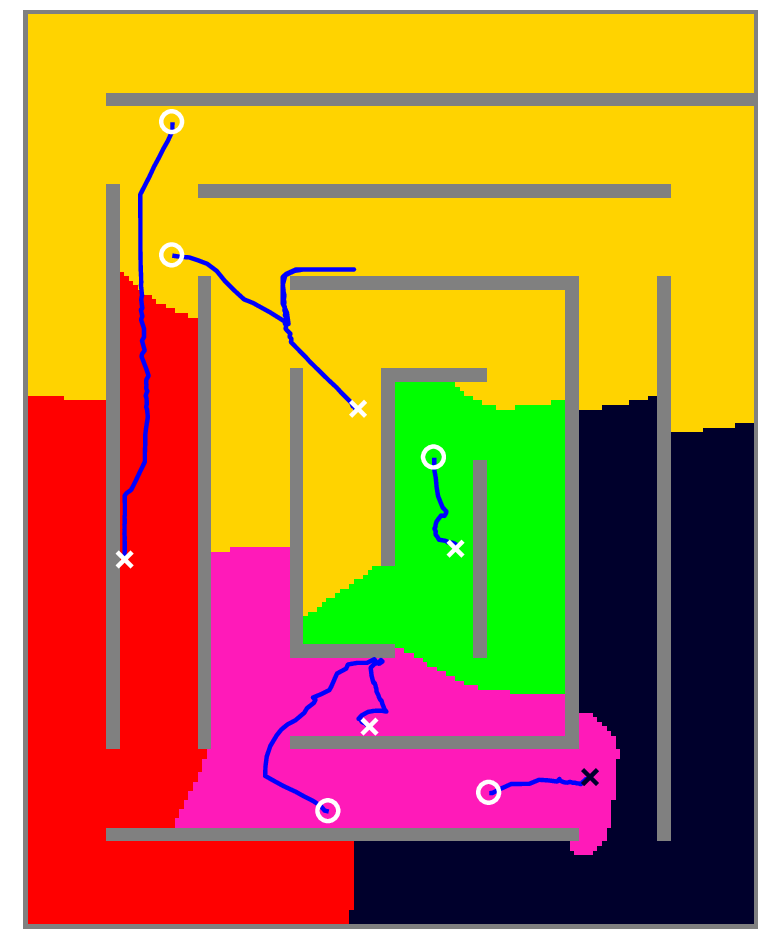}
                \label{Subfig:MoveSpiral}
                }
                \hfil
        \subfloat[Maze.]{ \includegraphics[width=0.23\linewidth]{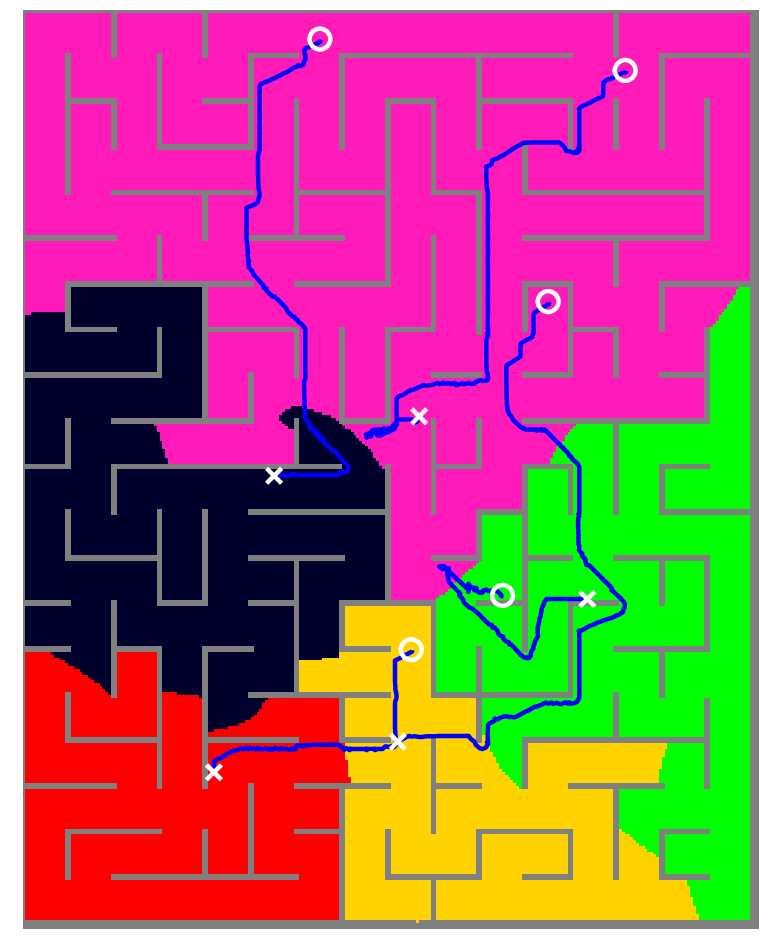}
                \label{Subfig:MoveMaze}
                }
    \caption{Example of partitioning in the four different environments under control law~\eqref{Eq:ControlLaw1} (top row) and under control law~\eqref{Eq:ControlLaw2} (bottom row). The resulting partitions  are shown in different colors. The initial locations of the generators are represented with white circumferences and the final location with white crosses. Blue lines in the bottom row represent the path of the generators until the equitable partitions are achieved. In the cases where a generator ended outside its own partition, it has been plotted with the color of its partition instead of white to show the correspondences.}
    \label{Fig:Example}
\end{figure*}

In this section we present simulation results for the partitioning algorithms and the complete approach to the persistent coverage problem.
For the partitioning, we consider four different, non-convex environments of 10x8m and increasing complexity. We call them \emph{Open Rooms, Rooms, Spiral} and \emph{Maze}, respectively. They can be seen in Fig.~\ref{Fig:Example}.
These simulations have been carried out with $N=5$ robots of radius $r_i = 0.1 m$, a circular coverage area $\Omega_i$ of radius $r_i^{cov} = 0.2m$, a production $\alpha_i = 1, \, \forall \, \mathbf{q} \in \Omega_i,$ and $\rho_i^{\max}= 20$.
The objective, decay and importance of the coverage are
\begin{gather}
    Z^* = 80 + 20 \, \frac{\mathbf{q}_y}{|\mathcal{Q}|_y},\\
    d = 0.9995, \label{Eq:Decay} \\
    \phi = 0.5 + 0.5 \, \frac{\mathbf{q}_y}{|\mathcal{Q}|_y},
\end{gather}
where $\mathbf{q}_y$ are the $y$-coordinates of point $\mathbf{q}$ and $|\mathcal{Q}|_y$ is the $y$-size of the environment $\mathcal{Q}$.
The resulting work function $\lambda(\mathbf{q})$ can be seen in Fig.~\ref{Fig:Workmap} for the \emph{Rooms} environment, normalized by $\int_\mathcal{Q} \lambda(\mathbf{q}) \, d\mathbf{q}$ and multiplied by 100.
The upper part of the environment requires less work because of its lower importance and objective and the central and lower parts require more work due to the peak of the decay in the center and the higher importance and objective at the bottom.

The positive constant of the saturation function~\eqref{Eq:Saturation} is set to $k_{sat} = 3$ and the gains of the control law are different for the different environments and are updated online depending on the value of $\partial H / \partial w_i$. In particular, $k_g$ takes the values from Table~\ref{Tab:kg} and $k_w$ is assigned according to
\begin{equation}
    k_w = \left\{ 
    \begin{aligned}
        & k_w^1, && \text{if } |\partial H / \partial w_i| > \beta_1, \\
        & k_w^2, && \text{if } \beta_1 \leq |\partial H / \partial w_i| < \beta_2, \\
        & k_w^3, && \text{if } \beta_2 \leq |\partial H / \partial w_i| < \alpha_3, \\
        & k_w^4, && \text{if } |\partial H / \partial w_i| \leq \beta_3,
    \end{aligned}
    \right.
\label{Eq:kw}
\end{equation}
with the values of the constants presented in Table~\ref{Tab:kw}.
These parameters were manually tuned in order to speed up the convergence of the partitions. In practice, there is no need to change these parameters for different environments.
On the downside, the convergence time of the equitable partitions may vary and affect some environments more than others if they are not appropriately tuned. In any case, this time is negligible when compared with the duration of the task, that can be infinite.

In our simulations the map is represented by a matrix of pixels (an image), so, even when the partition method is continuous, there is some degree of discretization.
To compute the gradients, we keep track of the partition associated to each pixel, obtaining the boundaries by checking if all the neighbor pixels belong to the same one.
Then, for the pixels in the boundary we compute the geodesic distance and $(\mathbf{q} - h^1_{\mathbf{q},\mathbf{g}_j})$ in Eq.~\eqref{Eq:df_dxk}.
From there, we can compute Eq.~\eqref{Eq:invn} and the gradient in Eq.~\eqref{Eq:dH_dwi} by summing this value for all the points of the partition boundary.
Although this procedure looks complex, it is not computationally expensive, requiring in our case 31ms per iteration and robot.

\begin{table}[!ht]
\small\sf\centering
\caption{Values for $k_g$ depending on the environment type.}
\centering
\begin{tabular}{cccc}
\toprule 
\emph{Open Rooms} & \emph{Rooms} & \emph{Spiral} & \emph{Maze} \\
\midrule 
$5$ & $5$ & $2$ & $0.1$ \\
\bottomrule 
\end{tabular}
\label{Tab:kg}
\end{table}

\begin{table}[!ht]
\small\sf\centering
\caption{Values for $k_w$ depending on the environment.}
\begin{tabular}{ccccc}
\toprule
& \emph{Open Rooms} & \emph{Rooms} & \emph{Spiral} & \emph{Maze} \\
\midrule
$k_w^1$ & $5$ & $5$ & $15$ & $15$ \\
$k_w^2$ & $100$ & $100$ & $75$ & $150$ \\
$k_w^3$ & $500$ & $500$ & $375$ & $1500$ \\
$k_w^4$ & $5000$ & $5000$ & $1500$ & $15000$ \\
$\beta_1$ & $5\cdot10^{-3}$ & $5\cdot10^{-3}$ & $5\cdot10^{-3}$ & $10^{-2}$ \\
$\beta_2$ & $3\cdot10^{-4}$ & $2\cdot10^{-4}$ & $5\cdot10^{-4}$ & $10^{-3}$ \\
$\beta_3$ & $10^{-5}$ & $10^{-5}$ & $5\cdot10^{-5}$ & $10^{-4}$ \\ 
\bottomrule 
\end{tabular}
\label{Tab:kw}
\end{table}

\subsection{Partitioning Example}

In the first place we present an example of the partitioning algorithms in the different environments.
In the top row of Fig.~\ref{Fig:Example} we show the resulting equitable partitions under~\eqref{Eq:ControlLaw1}, i.e., when only the weights are modified and the positions of the generators remain fixed. Although in some cases such as Fig.~\ref{Subfig:NoMoveOpenRooms} 
the resulting partitions are connected, it can be seen that in others they are not, see e.g., yellow partition in Fig.~\ref{Subfig:NoMoveMaze}.
These disconnections are avoided under~\eqref{Eq:ControlLaw2} as can be seen in the bottom row of Fig.~\ref{Fig:Example}. In this case, moving the generators towards the centroid of the connected components with the highest workload at the same time as the weights are changing, allows the robots to find equitable and connected partitions.
\begin{figure}[!ht]
        \centering
        \includegraphics[width=0.95\columnwidth]{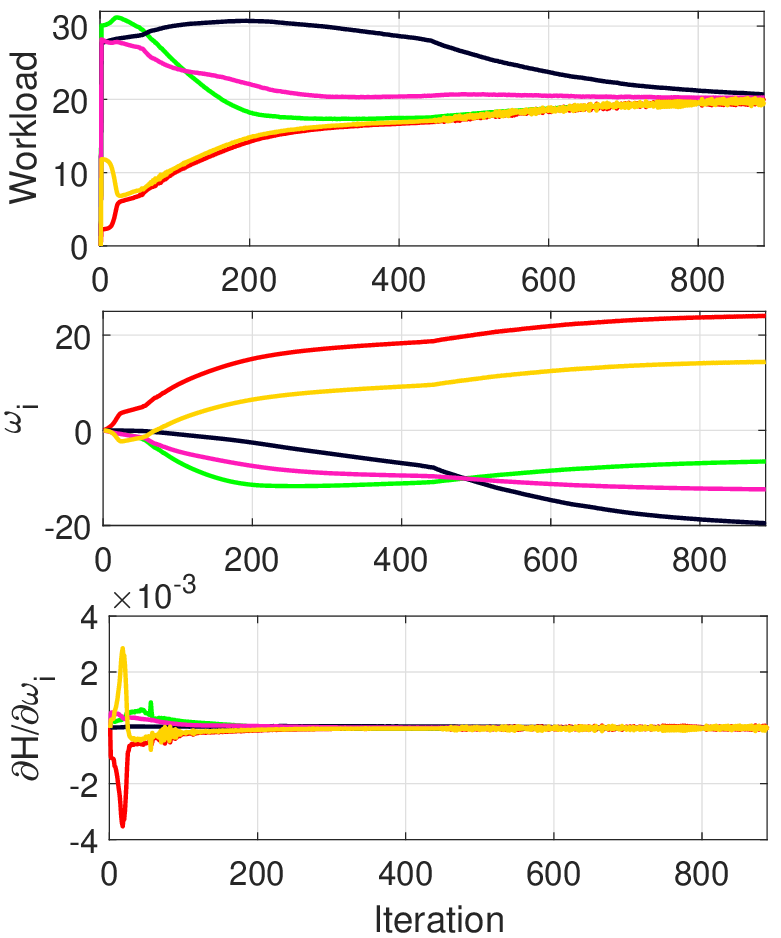}
                \label{Subfig:NoMoveWorkload}
    \caption{Evolution of the workload, weight and gradient of the cost function with respect to the weight for the \emph{Spiral} environment under control law~\eqref{Eq:ControlLaw1}. Different colors represent the variables for the different robots.}
    \label{Fig:ExampleNoMoveData}
\end{figure}

Now we focus in the particular example of the \emph{Spiral} environment to assess the evolution and convergence under both~\eqref{Eq:ControlLaw1} and~\eqref{Eq:ControlLaw2}.
In Fig.~\ref{Fig:ExampleNoMoveData} we represent the evolution of the workload, the weights and the gradient of the cost function with respect to the weight for each robot in a different color under~\eqref{Eq:ControlLaw1}.
First of all, in the top Figure we can observe that all the workloads converge to the same percentage value ($20\%$), that is, to the equity.
Convergence is assumed when the maximum difference between the workload of the partitions is lower than a 5\%.
Secondly, it is noteworthy that the weights associated to the partitions at the bottom of the map (black and magenta) are negative, while the ones at the top (red and yellow) are positive. This makes sense when we consider that the workload is distributed as in Fig.~\ref{Fig:Workmap}. Since the upper part is less important, robots in that area are assigned larger partitions, and therefore positive weights.
Finally, the value of the gradient is represented in the bottom plot.

\begin{figure}[!ht]
        \centering
        \includegraphics[width=0.45\textwidth]{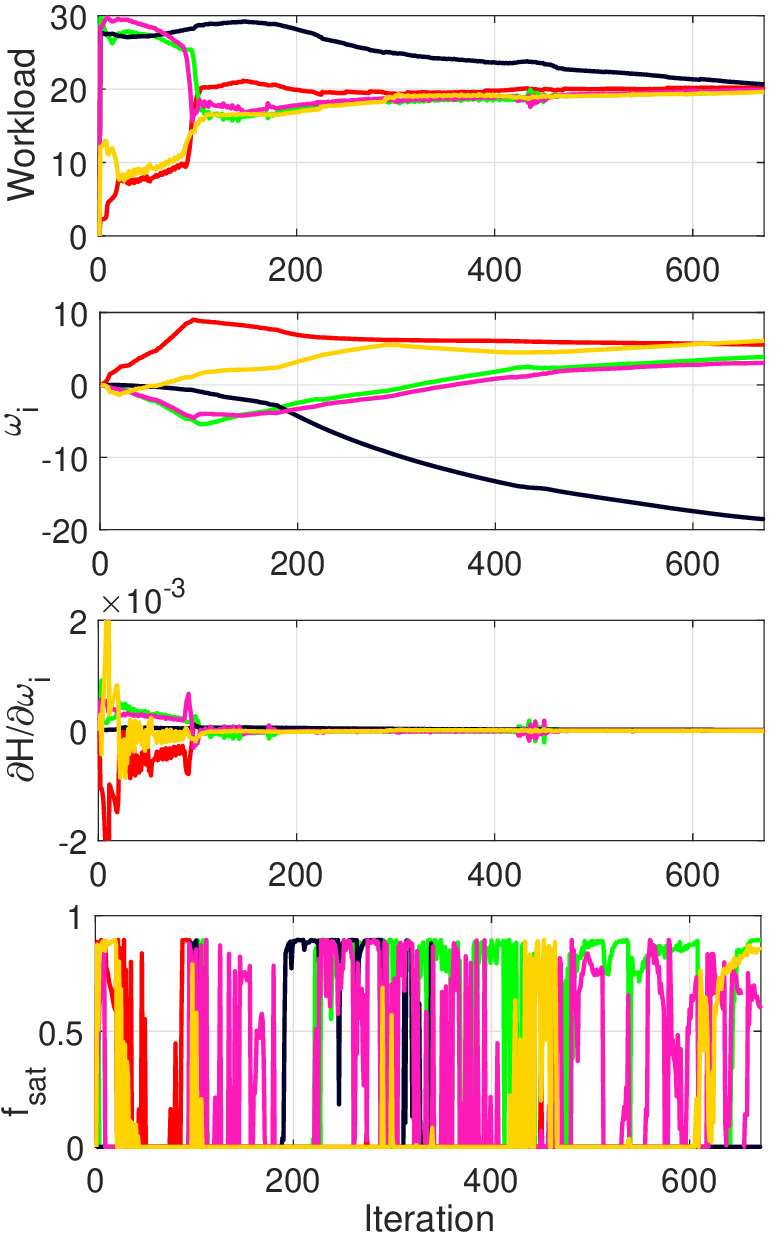}
    \caption{Evolution of the workload, weight, gradient of the cost function with respect to the weight, and saturation function for the \emph{Spiral} environment under control law~\eqref{Eq:ControlLaw2}. Different colors represent the variables for the different robots.}
    \label{Fig:ExampleMoveData}
\end{figure}

Under~\eqref{Eq:ControlLaw2}, in the beginning the weights and workloads also evolve slowly as shown in Fig.~\ref{Fig:ExampleMoveData}. 
The convergence speed of this law is similar to that in~\eqref{Eq:ControlLaw1}.
The main difference appears in the bottom plot of the figure, where we show how often the motion of the generators is saturated, $f_{sat}$ in Eq.~\eqref{Eq:Saturation}.
Values equal to zero mean that moving the generator goes against obtaining equity. 
The main consequence is that the evolution of the partitions is not as smooth as with~\eqref{Eq:ControlLaw1} but all the resulting partitions are connected.
In the following section we carry out a parametric analysis to be able to generalize the results.

\subsection{Parametric Analysis of the Partitioning Algorithms}

We analyze now the evolution of the performance of the partitioning algorithm when the parameters of the problem change. In particular, we performed a Monte Carlo analysis of 10 runs for different initial positions of the generators and different work functions in all the environments. In order to represent the results concisely, we refer to the different environments by their initials as follows: \emph{Open Rooms} as OR, \emph{Rooms} as R, \emph{Spiral} as S and \emph{Maze} as M.
The 10 initial positions of the generators were selected randomly and the variation of the work function was done through the decay function. In~\eqref{Eq:Decay} the decay is set to be a 2D Gaussian centered at $|\mathcal{Q}|_x/2$ and $|\mathcal{Q}|_y/2$, i.e., in the middle of the environment. Instead of this center, we chose a random point for each trial.
We execute this analysis for~\eqref{Eq:ControlLaw1} and~\eqref{Eq:ControlLaw2} and refer to them as W and WG respectively, representing that only the power weights or the power weights and the generator positions are modified.

\begin{figure}[!ht]
    \centering
    \subfloat[Different initial positions.]{
        \includegraphics[width=0.47\textwidth]{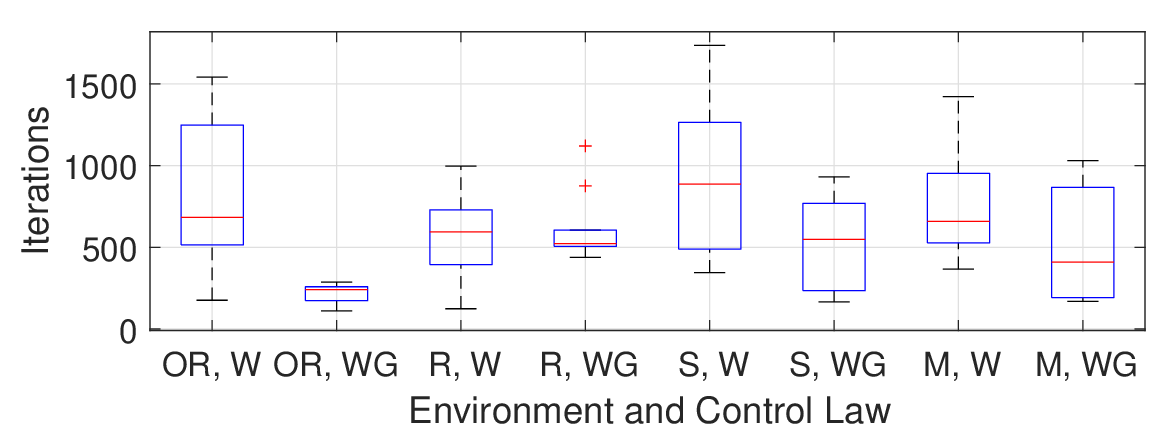}
        \label{Subfig:NItPositions}
            }
            
    \subfloat[Different decay functions.]{
        \includegraphics[width=0.47\textwidth]{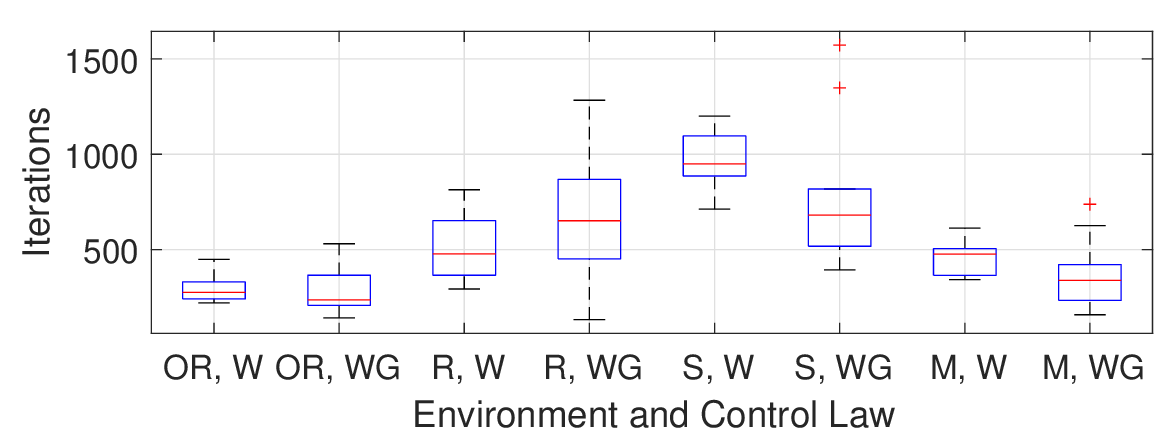}
        \label{Subfig:NItDecays}
            }
    \caption{Boxplot of the number of iterations required to converge to the equitable partition.}
    \label{Fig:NIt}
\end{figure}

In the first place we show in Fig.~\ref{Fig:NIt} two boxplots of the number of iterations required for the algorithm to converge to the equitable partition. Fig.~\ref{Subfig:NItPositions} shows the results for different initial positions of the generators and Fig.~\ref{Subfig:NItDecays}, for different decay functions.
The most important feature of these results is that, for the same environment, WG converges in general faster than W.
Among the different environments, the speed of convergence depends on the complexity of the environment and on the tuning of the gains.

In the second place we pay attention to the connectivity of the partitions. Table~\ref{Tab:DiscPart} gathers the number of partitions that resulted disconnected out of the 50 regions corresponding to each cell. As expected, this number is drastically reduced with the WG control law. In fact, for \emph{Open Rooms} and \emph{Rooms}, it converges to connected partitions in all cases. However, for \emph{Spiral} and \emph{Maze} the partitions result disconnected in some cases.

\begin{table}[!ht]
\caption{Number of disconnected partitions.\label{Tab:DiscPart}}
\centering
\begin{tabular}{ccccc}
\cline{2-5}
& \multicolumn{2}{c}{Different Positions} & \multicolumn{2}{c}{Different Decays} \\
\cline{2-5}
& W & WG & W & WG \\
\hline
\emph{Open Rooms} & 1 & 0 & 4 & 0 \\
\emph{Rooms} & 30 & 0 & 29 & 4 \\
\emph{Spiral} & 12 & 4 & 26 & 9 \\
\emph{Maze} & 30 & 14 & 23 & 11 \\
\hline 
\end{tabular}

\end{table}

\begin{figure}[!ht]
    \centering
    \includegraphics[width=0.48\textwidth]{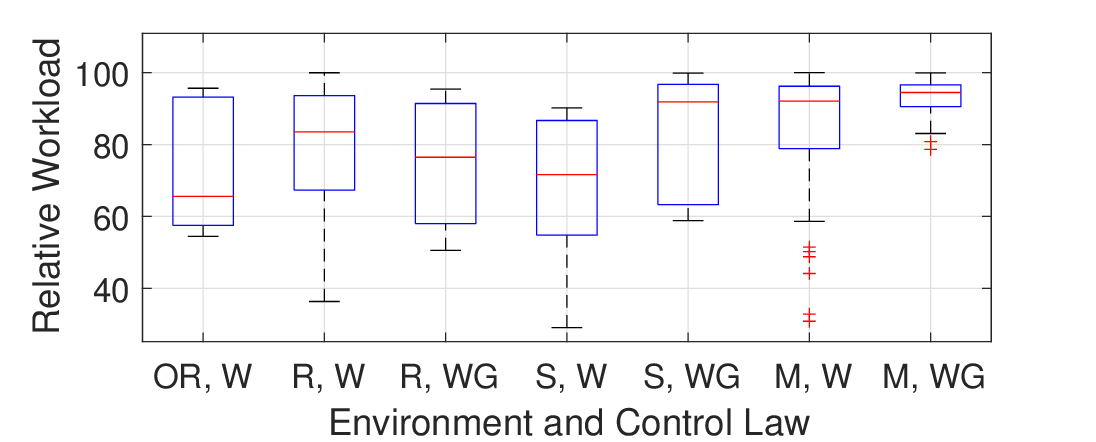}
    \caption{Boxplot of the relative workload of the biggest connected component with respect to the workload of the entire partition.}
    \label{Fig:RelWorkload}
\end{figure}

To evaluate the nature and the quality of the still non-connected partitions in Fig.~\ref{Fig:RelWorkload} we represent in percentage the relative workload of the biggest connected component with respect to the workload of the entire partition.
One can see that under WG, this value is on average above the 80\%.
This means that almost all the work of the partition is on the biggest component, whose centroid was followed by the generator. Therefore, a simple reassignment of the smaller disconnected regions could be done to reach total connectivity with only a small deviation from the equitable partitioning.

\subsection{Coverage Planning Results}

\begin{figure}[!ht]
    \centering
    \includegraphics[width=0.25\textwidth]{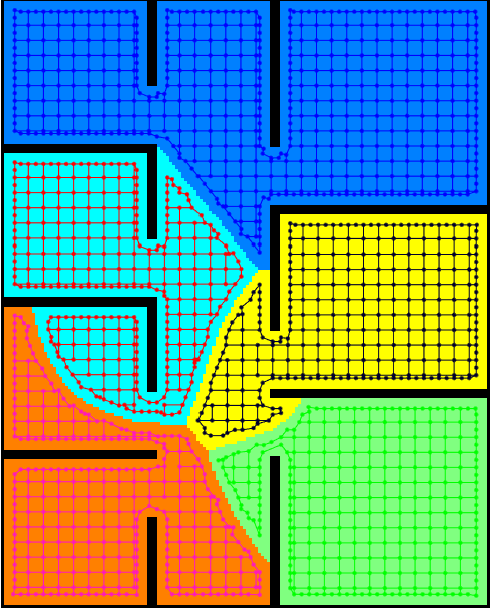}
    \caption{Example of partitions and path graphs for the \emph{Rooms} environment.}
    \label{Fig:ExampleGraph}
\end{figure}

In this part of the simulations we analyze the coverage planning solution and the performance of the entire approach with an example.
Fig.~\ref{Fig:ExampleGraph} shows the partition of the \emph{Rooms} environment and the path graphs of the robots. According to the work map showed in Fig.~\ref{Fig:Workmap}, the robot assigned to the blue partition has to cover a bigger area than the other four. These other four share the peak of the work map and each of them is additionally in charge of covering a big room on the right-hand side or one and a half small rooms on the left-hand side.

\begin{figure*}[!ht]
    \centering
    \subfloat[k = 60.]{
        \includegraphics[height=4.7cm]{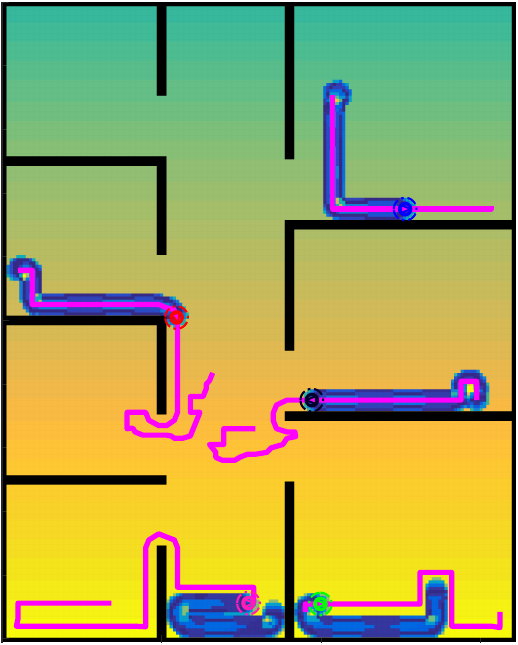}
        \label{Subfig:Map1}
            }
            \hfil
    \subfloat[k = 275]{
        \includegraphics[height=4.7cm]{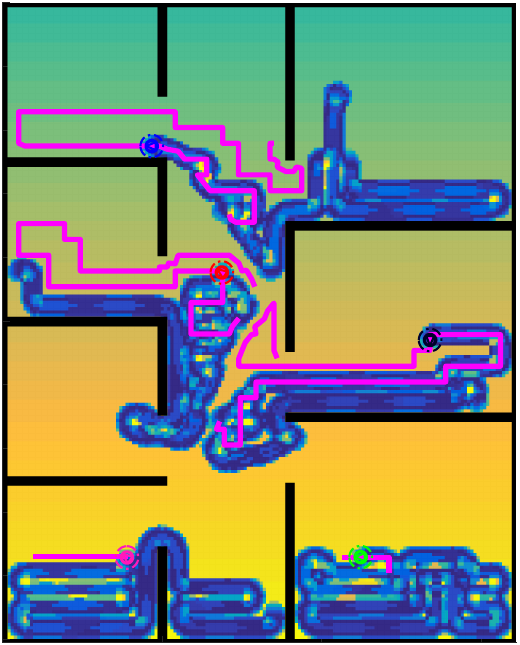}
        \label{Subfig:Map2}
            }
            \hfil
    \subfloat[k = 400.]{
        \includegraphics[height=4.7cm]{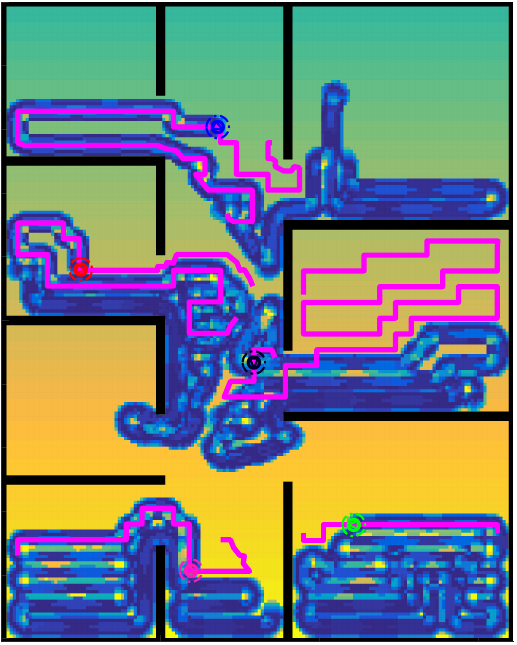}
        \label{Subfig:Map3}
            }
            \hfil
    \subfloat[k = 500]{
        \includegraphics[height=4.7cm]{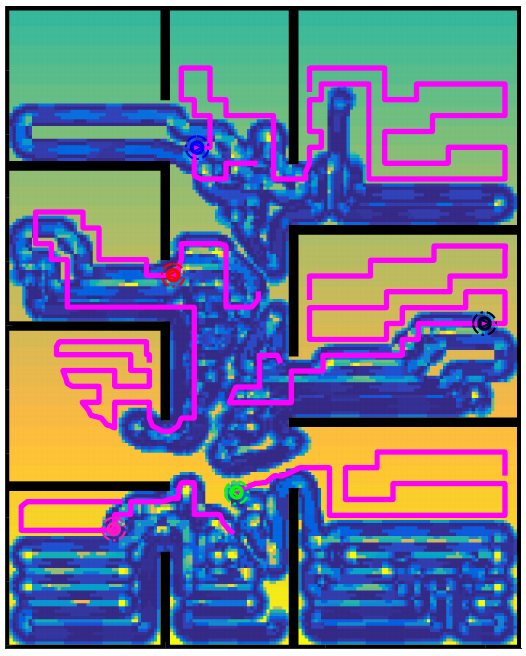}  
        \label{Subfig:Map4}
            }
            
    \subfloat[k = 1150.]{
        \includegraphics[height=4.7cm]{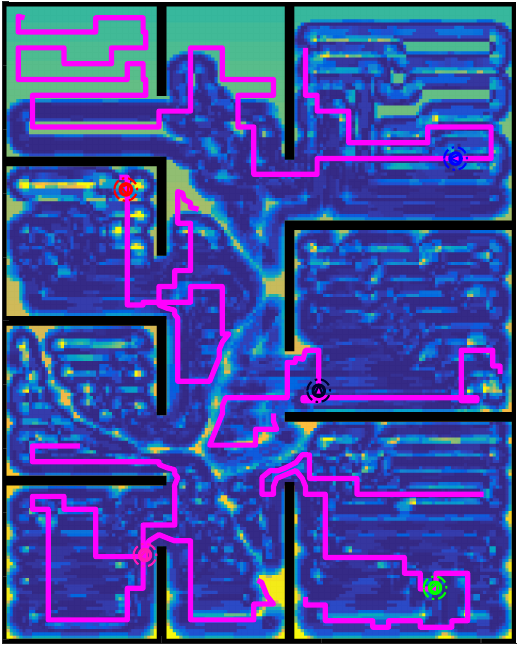}
        \label{Subfig:Map5}
            }
            \hfil
    \subfloat[k = 1700.]{
        \includegraphics[height=4.7cm]{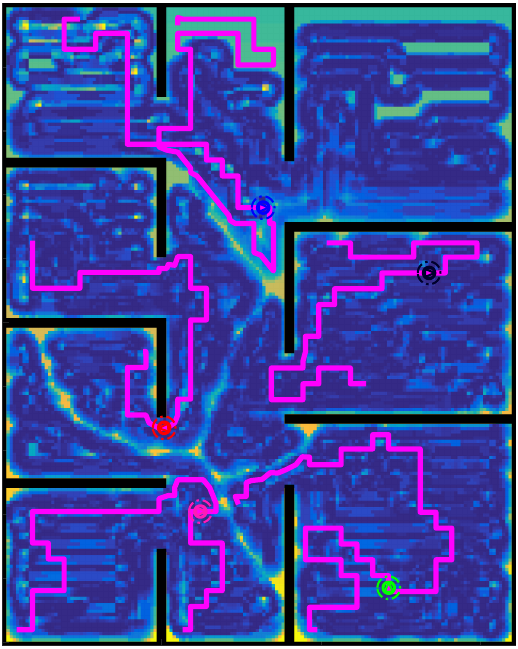}
        \label{Subfig:Map6}
            }
            \hfil
    \subfloat[k = 2000.]{
        \includegraphics[height=4.7cm]{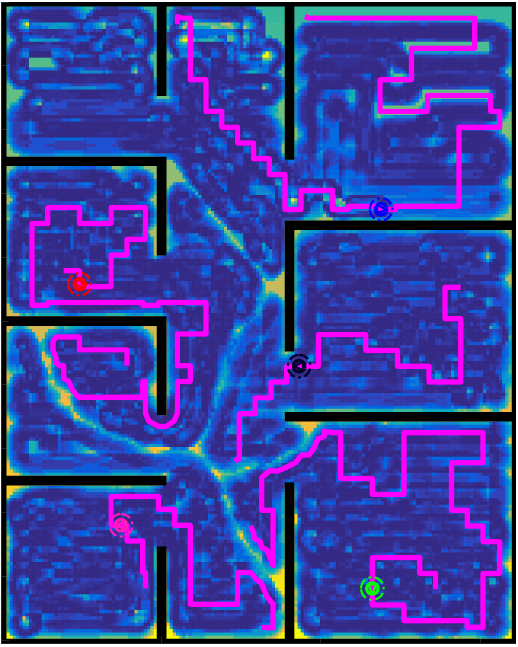}
        \label{Subfig:Map7}
            }
            \hfil
    \subfloat[k = 2500.]{
        \includegraphics[height=4.7cm]{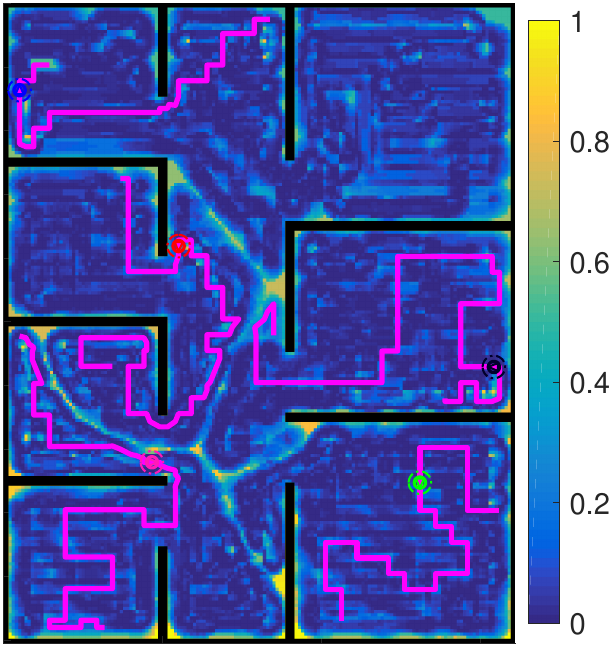}  
        \label{Subfig:Map8}
            }
    \caption{Example of coverage evolution. The background color map represents the value of $\varepsilon(k)$. The robots are depicted in different colors with their coverage areas as dash dotted circumferences. Magenta lines show the current paths of the robots.}
    \label{Fig:Map}
\end{figure*}

In order to evaluate the quality of the paths and the provided coverage we normalize the coverage error at a single point at time $k$,
\begin{equation}
    \varepsilon(k) \equiv \varepsilon(\mathbf{q},k) = \frac{e(k)}{{Z^*}^2},
\end{equation}
and calculate the mean normalized error over the environment as
\begin{equation}
\label{Eq:normalizedError}
    \overline{\varepsilon}(k) =  \frac{\int_{\mathcal{Q}_f} \varepsilon(k) d\mathbf{q}}{|\mathcal{Q}_f|}.
\end{equation}

In Fig.~\ref{Fig:Map} we show the coverage error $e(\mathbf{q},k)$ of the environment for eight different time instants. In Fig.~\ref{Subfig:Map1} the robots are starting to cover the environment and, therefore, the coverage error is initially proportional to the importance of the points and their objective.
The paths of the top row, Fig.~\ref{Subfig:Map1}-\ref{Subfig:Map4}, represented in magenta, cover the partitions from bottom to top in a sweep-like manner. This demonstrates that the paths go through the points with the highest coverage error.
In Fig.~\ref{Subfig:Map5}, almost all the robots have already covered their partitions for the first time.
The blue robot needs a little more time since it has the biggest partition. 
The paths of the robots in Fig.~\ref{Subfig:Map5} and~\ref{Subfig:Map6}, fill the gaps that they have previously left uncovered.
At this point, the coverage error is already small in the majority of the points.
Eventually, in the steady state, Fig.~\ref{Subfig:Map7}-\ref{Subfig:Map8}, the robots keep moving to maintain the coverage as close as possible to the objective, paying special attention to the zones where more work is required.

\begin{figure}[!ht]
    \centering
    \subfloat{
        \includegraphics[width=0.45\textwidth]{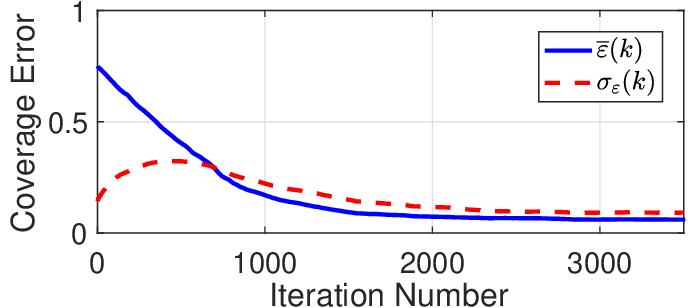}
        \label{Subfig:QuadCovError}
            }
            
    \subfloat{
        \includegraphics[width=0.45\textwidth]{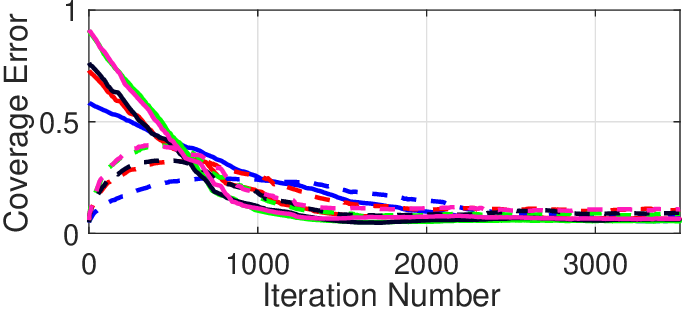}
        \label{Subfig:QuadCovErrorPart}
            }
    \caption{Evolution of the mean coverage error (solid lines) and its standard deviation (dashed lines). (Top) Complete environment. (Bottom) Separated by partitions.}
    \label{Fig:QuadCovError}
\end{figure}

The performance of our coverage strategy is assessed in Fig.~\ref{Subfig:QuadCovError}. We depict the mean coverage error and its standard deviation for the simulation. One can see that, at the beginning, when the environment is totally uncovered, the error is around 0.7 due to the importance function. After that, it decreases rapidly and converge to a small steady-state value. In fact, on average, the points have under 7\% of error. The standard deviation increases at the beginning because the undercovered points keep deteriorating and some others become slightly overcovered when the robots go over them. However, it also decreases to a small steady-state value around 0.09.

A similar tendency persists inside the partitions of the robots as shown in Fig.~\ref{Subfig:QuadCovErrorPart}, where $\overline{\varepsilon}(k)$ is calculated only within $\mathcal{P}_i$. Nevertheless, it is interesting to see the differences between the biggest partition, the blue one, and the smaller ones, the magenta and green ones.
The error at the beginning in the magenta and green ones is higher because the importance of the points inside them is higher, as opposed to the blue one.
On the other hand, the reduction of the error in those partitions is faster because they are smaller and the robots need less time to cover them completely. On the contrary, the blue robot needs more time to cover its region and, therefore, its error decreases slower.

\begin{figure}[!ht]
    \centering
    \includegraphics[width=0.45\textwidth]{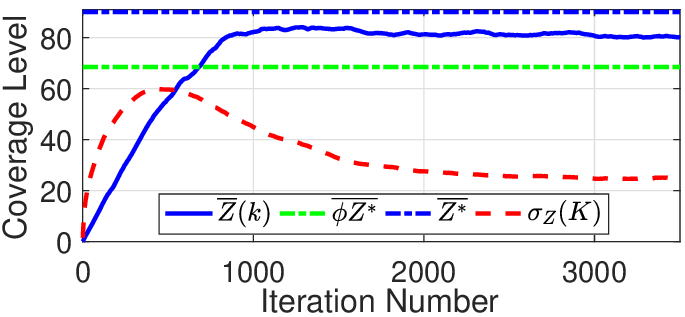}
    \caption{Evolution of the mean coverage level of the environment (blue line) with its standard deviation (red dashed line). The blue dash-dotted line represents the mean coverage objective and the green dash-dotted one, the mean coverage objective weighted by the importance of each point.}
    \label{Fig:CovLevel}
\end{figure}

Although the coverage error is the metric that defines the optimality of the coverage, it is difficult to visualize. A more intuitive representation is shown in Fig.~\ref{Fig:CovLevel}. The mean coverage level of the environment increases rapidly towards the mean coverage objective. However, its steady-state value is smaller due to the different importance of the points. This happens because the robots pay more attention to the most important points at the expense of leaving the least important worse covered. For this reason, the mean value lays between the mean coverage objective and the mean coverage objective weighted by the importance.

\begin{figure}[!ht]
    \centering
    \includegraphics[width=0.25\textwidth]{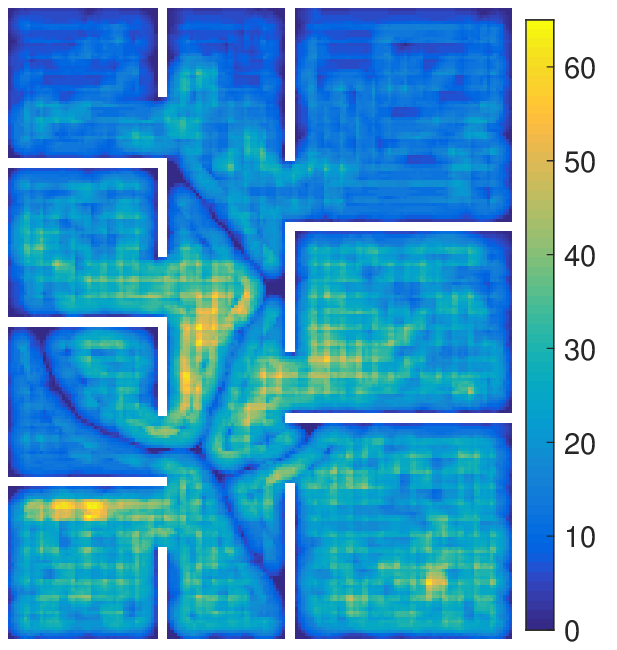}
    \caption{Number of times that the robots have covered each point of the environment.}
    \label{Fig:NCoverages}
\end{figure}

Finally, we show in Fig.~\ref{Fig:NCoverages} the number of times that the robots covered each point of the environment. The resemblance with the work map makes it clear that the robots visit more frequently the most important points and those with a lower decay, because the coverage deteriorates faster on them. The undercovered zones near the boundaries between partitions occur due to discretization in the simulation and due to the separation from the boundary to the boundary nodes of the graph.

\subsection{Comparison with centroidal Voronoi partitions}
To conclude the simulations, we compare the coverage results to those obtained using the Voronoi partition  algorithm described in~\cite{cortes2004coverage} to generate the partitions of the environment.
Both methods have been tested in the \emph{Rooms} environment, with the work function $\lambda$ represented in Fig.~\ref{Fig:Workmap}.
In the Voronoi partition we have used the workload as the density function to compute the centroid of the partition.

\begin{figure*}[!ht]
    \centering
    \begin{tabular}{cccc}
        \includegraphics[height=4cm]{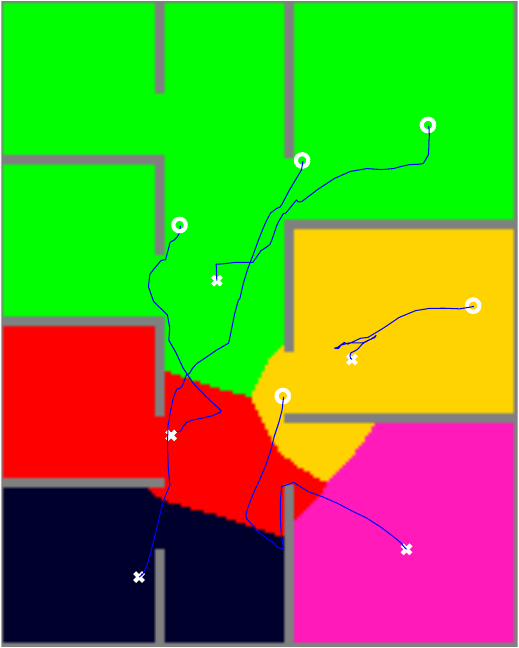} &
        \includegraphics[height=4cm]{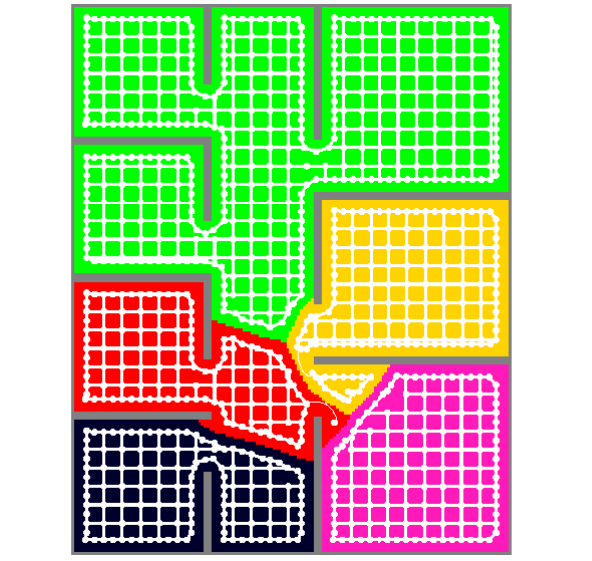} &
        \includegraphics[height=4cm]{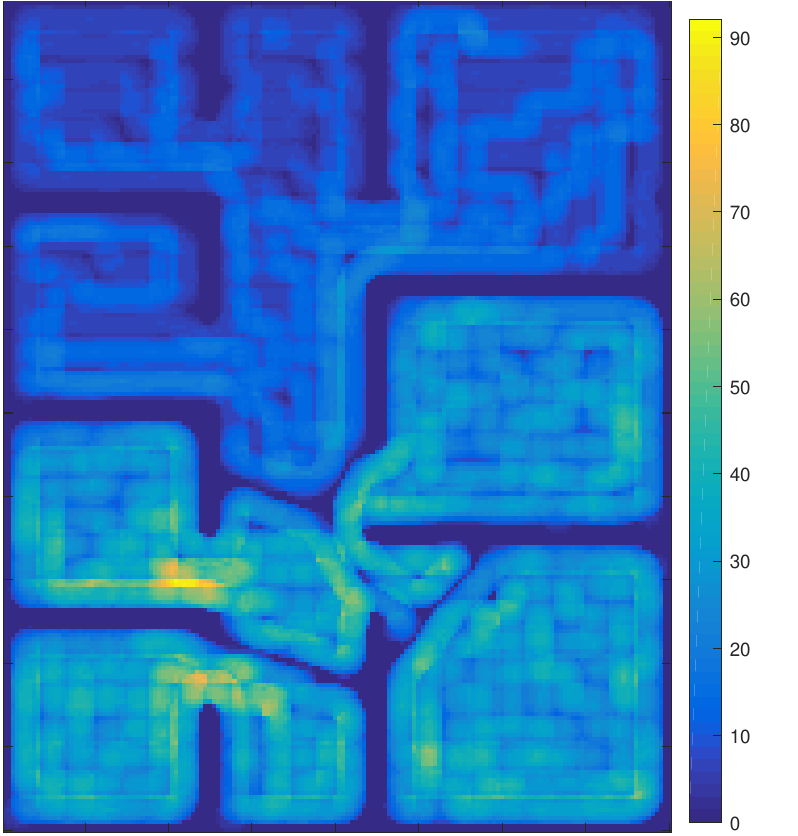} &
        \includegraphics[height=4cm]{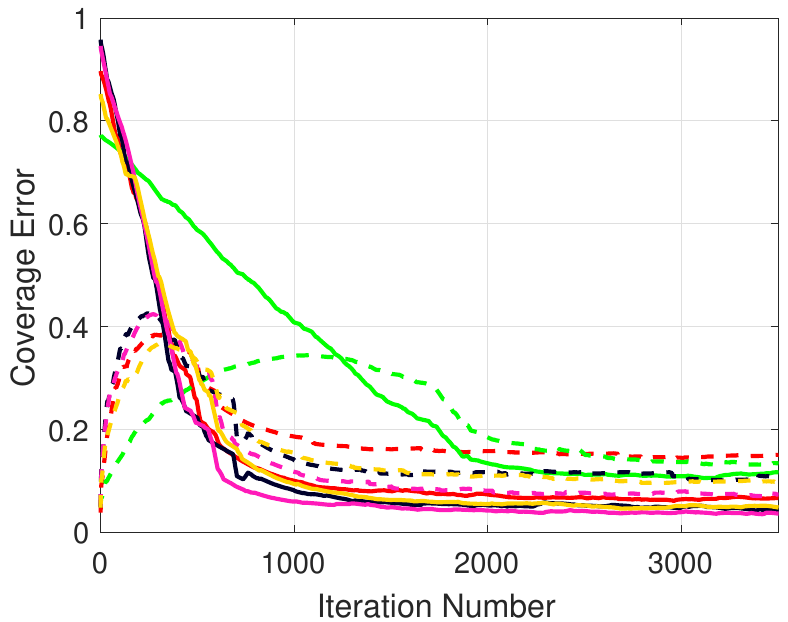}\\ 
        \includegraphics[height=4cm]{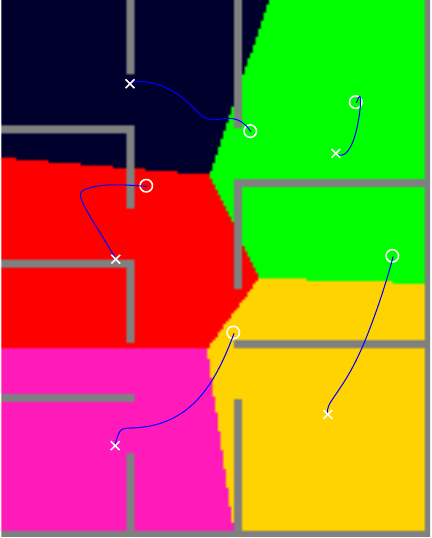} & 
        \includegraphics[height=4cm]{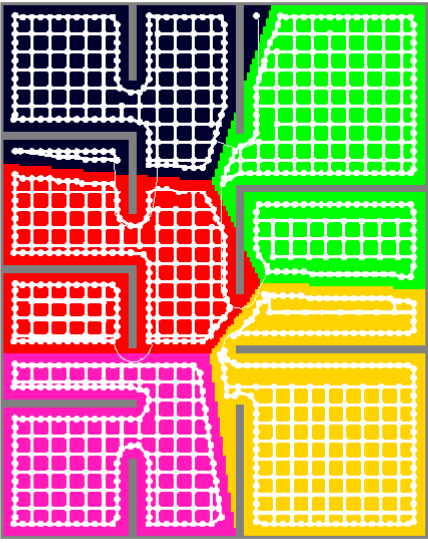} &
        \includegraphics[height=4cm]{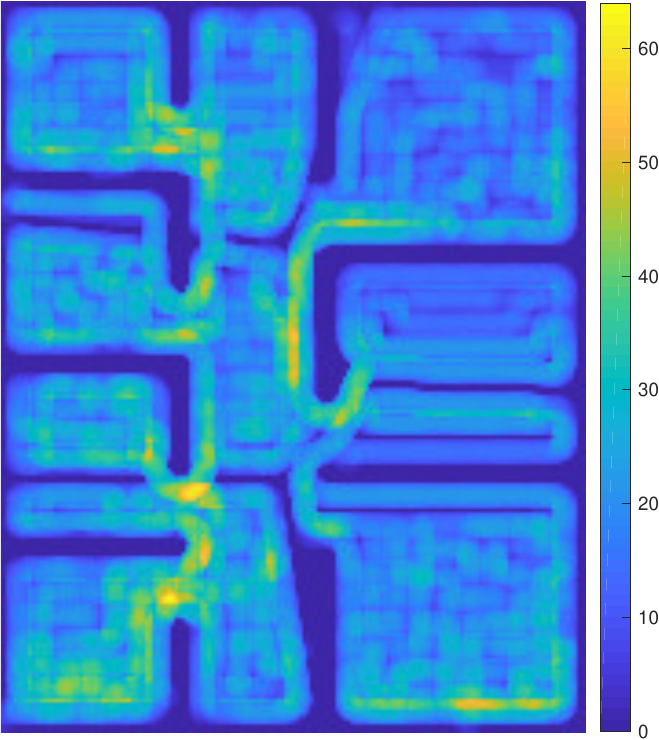} &
        \includegraphics[height=4cm]{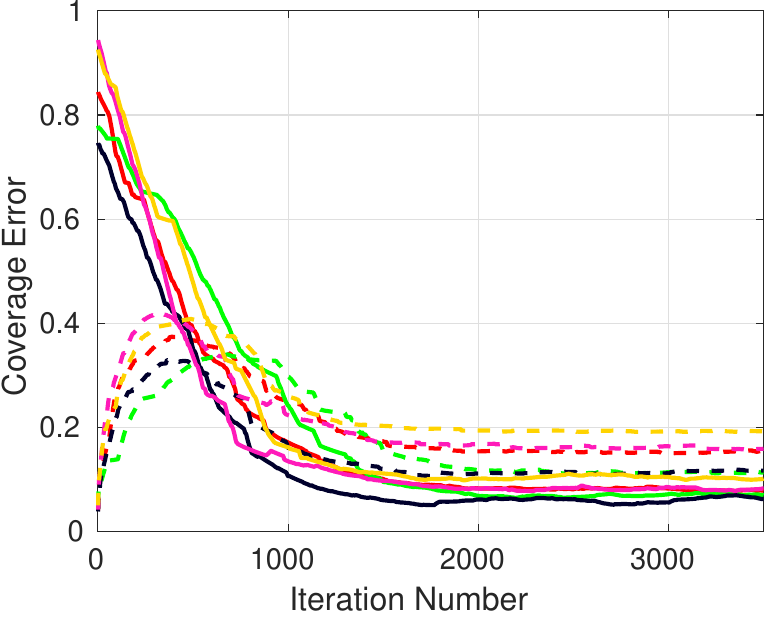}\\
        (a) Partitions & (b) Coverage graphs & (c) Number of Visits & (d) Coverage error 
    \end{tabular}
    \caption{Comparison of the solution with a standard Voronoi partition. The top row shows our approach and the bottom row shows the results with the Voronoi partition.} 
    \label{Fig:VoronoiComp}
\end{figure*}

In Fig.~\ref{Fig:VoronoiComp} (a) we show the partitions obtained with the two algorithms for one example of initial conditions of the generators.
In this particular example the yellow partition accounts for the $25.64\%$ of the total workload, whereas the black partition only accounts for $15.83\%,$ implying that first robot will need to do considerably more work than the second.
In addition, two out of the five partitions are not connected (black and green).
We have repeated this analysis for the 10 runs described in 
Table~\ref{Tab:DiscPart} for this environment to measure the number of disconnected partitions and the difference between the maximum and minimum workload. The Voronoi tessellation method has yielded 16 disconnected regions, as opposed to the zero regions obtained with our method. The mean difference of the 10 trials has been $10.3\%$ with a standard deviation of $1.36.$

In the following, we focus on the coverage quality when both sets of partitions are used together with the path planning algorithm described in Section~\ref{sec:coverage}, just for the partitions given in Fig.~\ref{Fig:VoronoiComp} (a).
In Fig.~\ref{Fig:VoronoiComp} (b) we show the coverage graphs and in Fig.~\ref{Fig:VoronoiComp} (c) we show the number of times that each node of the graphs was visited during the experiment.
In the Voronoi partitions, some robots invade other robots' partitions in order to move between the disconnected components\blue{, e.g., the green partition}.
More importantly, with the equitable partition, points at the bottom of the map are visited more often, accounting for their decay and importance in a better way.

Finally, in Fig.~\ref{Fig:VoronoiComp} (d) we show the normalized error (solid lines), eq.~\eqref{Eq:normalizedError}, for each partition and the standard deviation (dashed lines) computed over all the points in the integral.
The transient behavior in this case is similar for both partitions, with a slightly better performance, i.e., less settling time, when using the equitable partitions for those regions with high importance (every region but the green one).
In the steady state it can be seen that the equitable partition presents a smaller standard deviation, implying a more homogeneous coverage.

\section{Experimental Results}
\label{sec:experiments}

The final validation of our proposal was performed carrying out real-world experiments in a complex environment shown in Fig.~\ref{Subfig:ExpMap}. It is composed by a small room in the left side and a main room in the center that connects to a corridor through two different doors. The goal was to persistently cover the environment shown in the same figure in yellow, with $N=3$ TurtleBot II robots.
\footnote{We refer the reader to the accompanying video of this experiment for a complete visualization of the coverage solution.}

\begin{figure}[!ht]
        \centering
        \includegraphics[width=0.95\columnwidth]{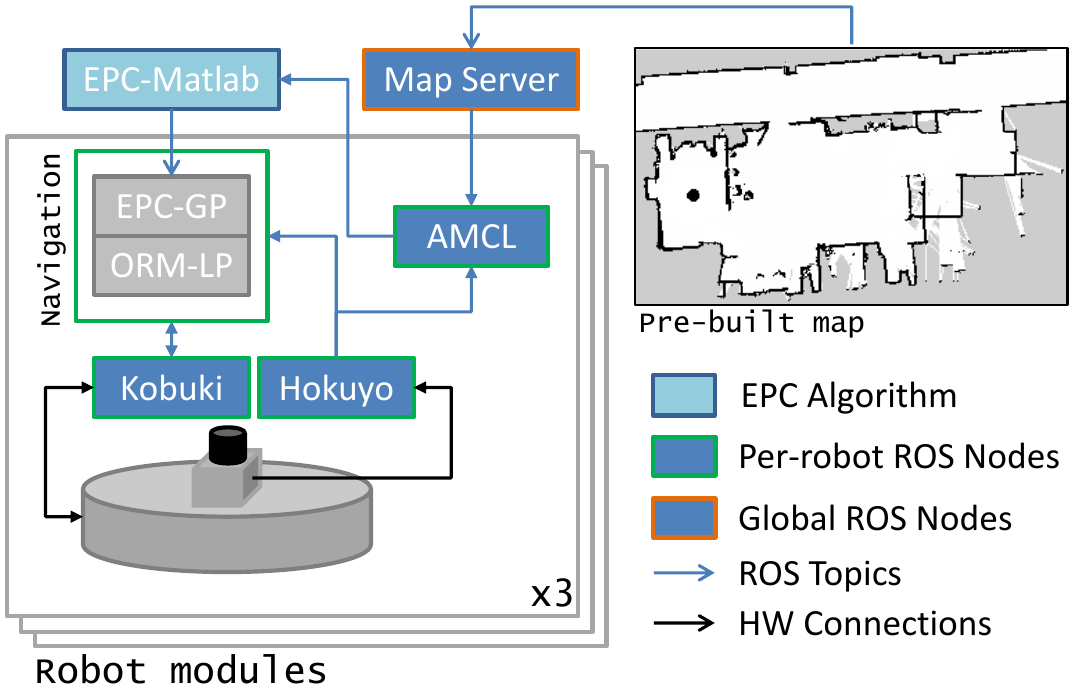}
        \caption{\label{fig:system-scheme} Multi-robot system architecture.}
\end{figure}

\subsection{Setup}
The multi-robot system was set up relying on the Robot Operating System (ROS) framework \cite{ros}. The robotics system was organized as shown in Fig. \ref{fig:system-scheme}. From the bottom up, two ROS nodes ---\textit{Kobuki} and \textit{Hokuyo}--- manage the TurtleBot II platform equipped with an on-board Intel NUC core i7 computer and a Hokuyo URG-04LX LiDAR sensor. These nodes provide odometry and laser readings to the \textit{AMCL} node and to the navigation stack.
The \textit{AMCL} node is in charge of localizing the robots on a pre-built map (obtained previously to the experiments) provided by the \textit{Map Server} node.
The navigation stack is, as usual, divided in two layers. At the bottom level, the \textit{ORM} \cite{orm} ---chosen because its efficacy in dense environments--- implements the basic obstacle avoidance capability.
On top of the \textit{ORM}, a dedicated global planner indicated as \textit{EPC-GP} was used. It wraps the Equitable Partition algorithm described in this paper, which is presently implemented in Matlab (\textit{EPC-Matlab}). The latter takes advantage of the Matlab robotics toolbox, which allows ROS communication between standalone ROS nodes and Matlab scripts. Specifically the \textit{EPC-Matlab} publishes the paths computed by the algorithm in a dedicated topic which is read by the \textit{EPC-GP}, that assumes them as global paths. Also, the \textit{EPC-Matlab} reads \textit{AMCL}-computed poses to update the coverage map. 
From the architectural point of view, the heavier computation was centralized in a single machine: only the \textit{Kobuki} and \textit{Hokuyo} nodes ran on the mobile platform computers while the other nodes ($3\times$AMCL, $3\times$Navigation and $1\times$Map Server) and the algorithm itself (\textit{EPC-Matlab}) resided in the base station connected with the robots through a WiFi router.

The coverage actions and values remained simulated in the experiment due to the complexity associated to measure these values in a real setup.
We defined a circular coverage area $\Omega_i$ of radius $r_i^{cov} = 0.186m,$ while the radius of the robots is equal to $r_i = 0.177 m$. The speed of the robots were fixed at $0.35~m/s$.
Their production and the objective functions were the same as in the simulations
and the importance of the coverage was
\begin{equation}
    \Phi = 0.7 + 0.3 \, \frac{\mathbf{q}_y}{|\mathcal{Q}|_y}.
\end{equation}

\subsection{Results}

\begin{figure}[!ht]
        \centering
        \subfloat[Environment map. The area to be covered is represented in yellow.]{
                \includegraphics[width=0.8\columnwidth]{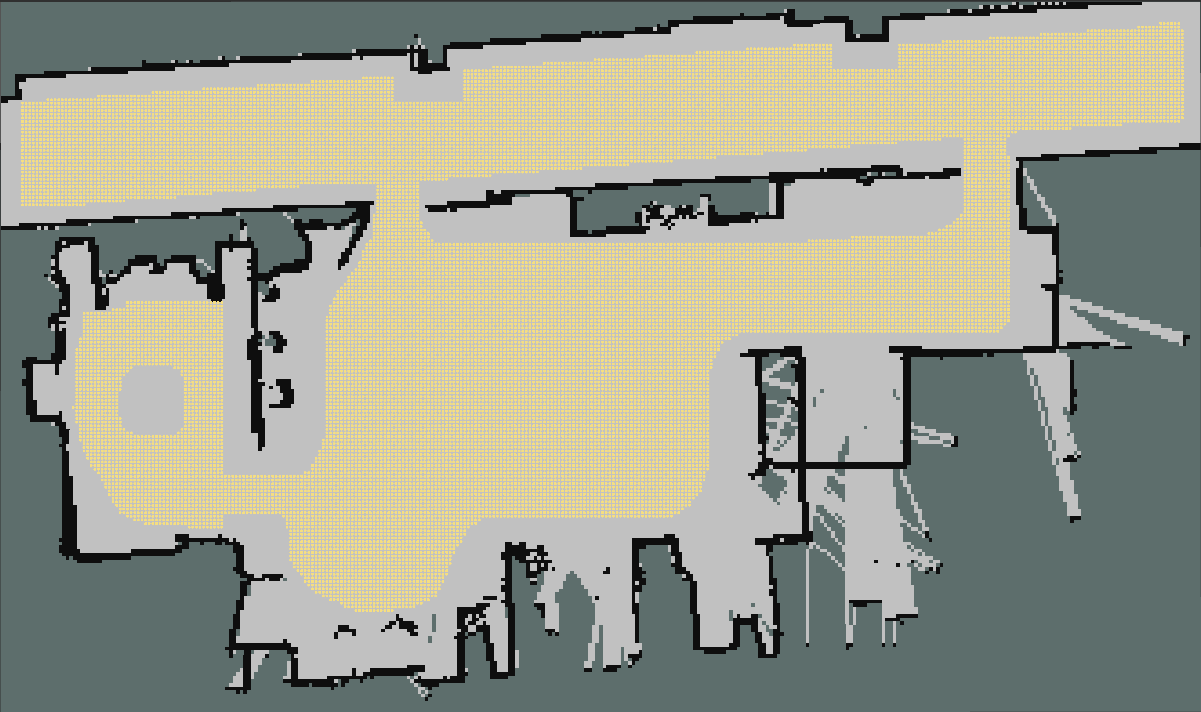}
                \label{Subfig:ExpMap}
                }
                
        \subfloat[Work map.]{
                \includegraphics[width=0.8\columnwidth]{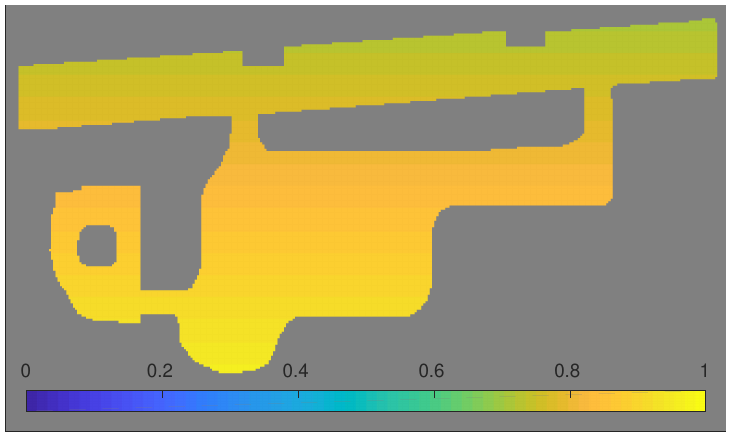}
                \label{Subfig:ExpImpMap}
                }
                
        \subfloat[Partitioned map with path graphs.]{
                \includegraphics[width=0.8\columnwidth]{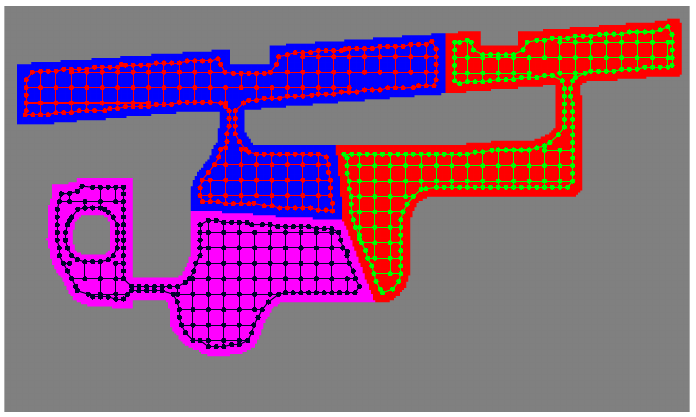}
                \label{Subfig:ExpPartitions}
                }
    \caption{Maps of the experimental environment and results from the partitioning and graph construction algorithms.}
    \label{Fig:ExpMaps}
\end{figure}

In this section we present two runs for different decays.
The decay was set to a uniform value of 0.9995 in experiment A and 0.995 in experiment B.
The resulting work for experiment A can be seen in Fig.~\ref{Subfig:ExpImpMap} and is proportional for experiment B. The lower part required more work while the corridor required less, due to its lower importance.
With these settings, the resulting partitions and path graphs computed by the robots were the ones depicted in Fig.~\ref{Subfig:ExpPartitions}. 
In both experiments they were the same since the only difference is a proportional decay.
The three partitions were equitable and self-connected with the three robots sharing the main room and two of them sharing the corridor through different doors.
The path graph of the each robot covers its entire partition and allows the robot to follow sweep-like paths.

\begin{figure*}[!ht]
        \centering
        \subfloat[Main room at $k=0$.]{
                \includegraphics[width=0.3\textwidth]{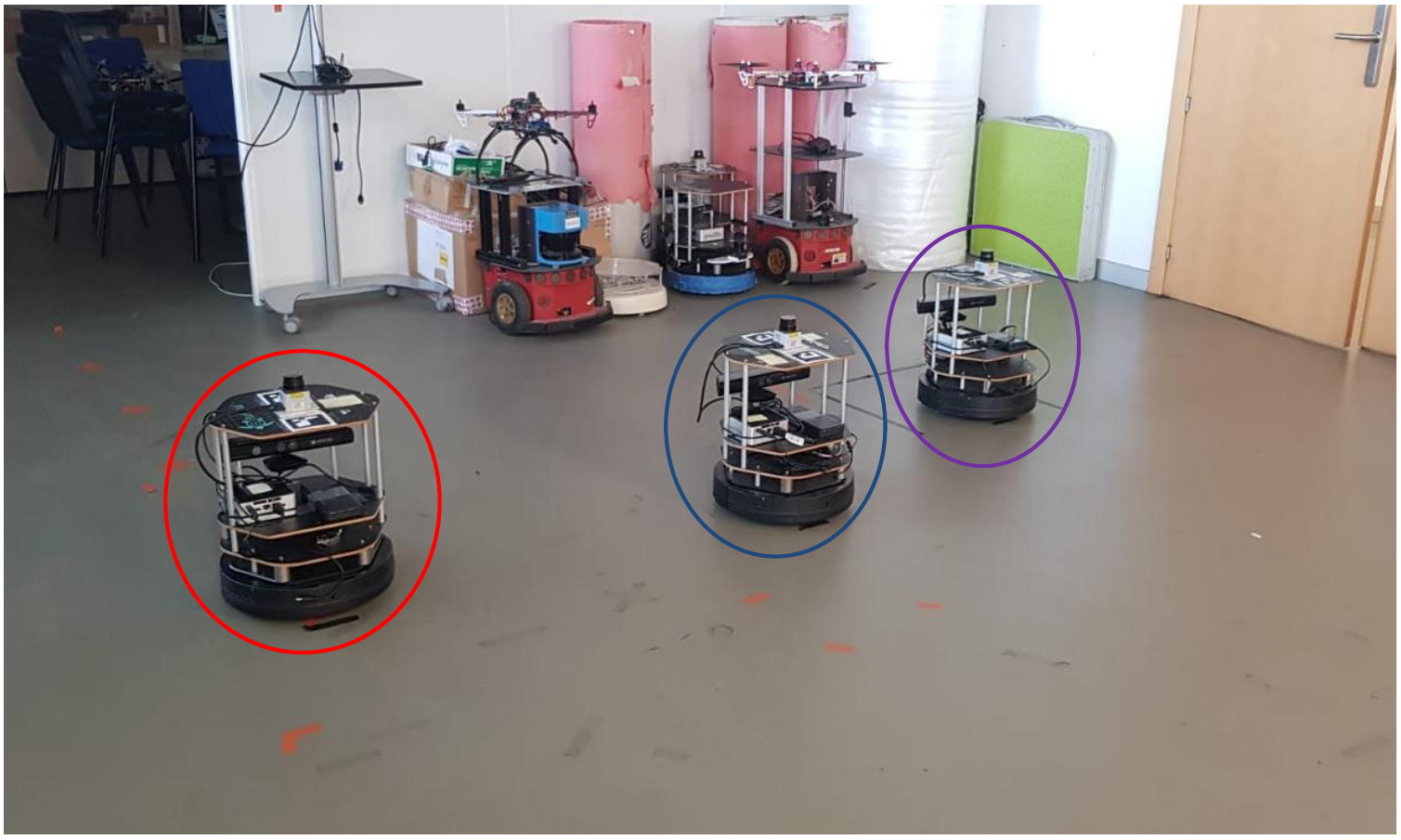}
                \label{Subfig:Snapshot0}
                }
                \hfil
        \subfloat[Main room at $k=100$.]{
                \includegraphics[width=0.3\textwidth]{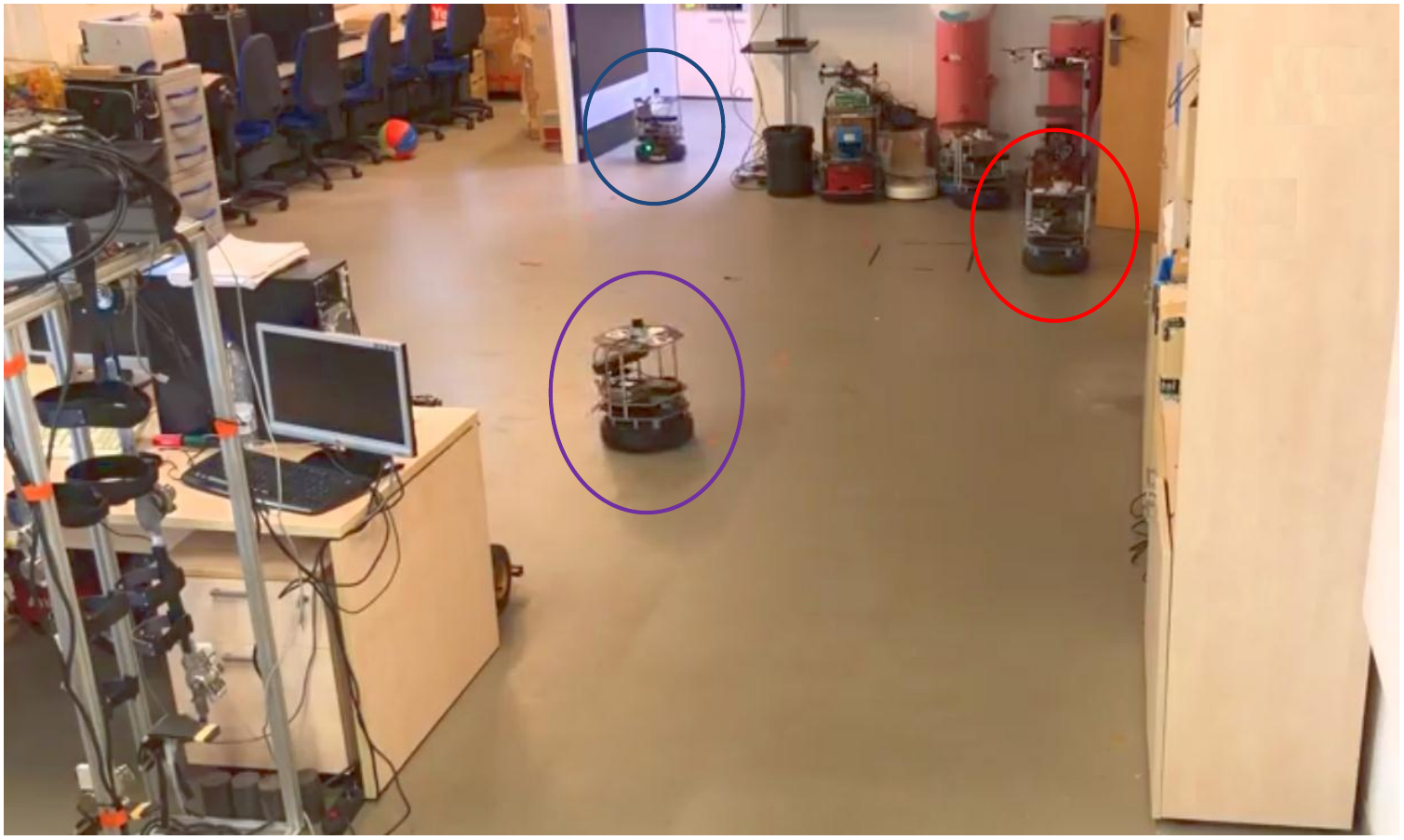}
                \label{Subfig:Snapshot1}
                }
                \hfil
        \subfloat[Coverage map and paths at $k=100$.]{
                \includegraphics[width=0.3\textwidth]{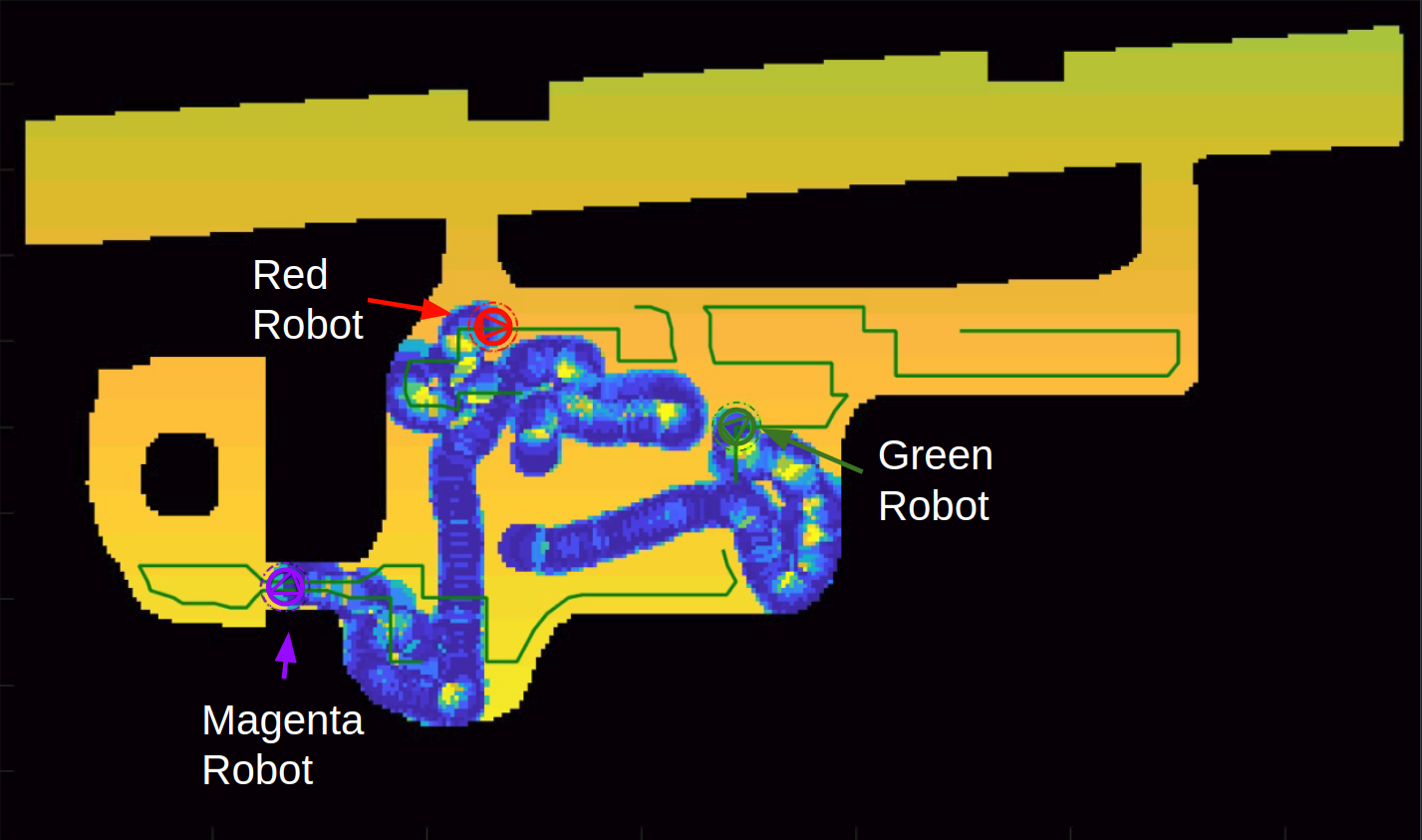}
                \label{Subfig:Snapshot1Matlab}
                }
                
        \subfloat[Corridor at $k=500$.]{
                \includegraphics[width=0.3\textwidth]{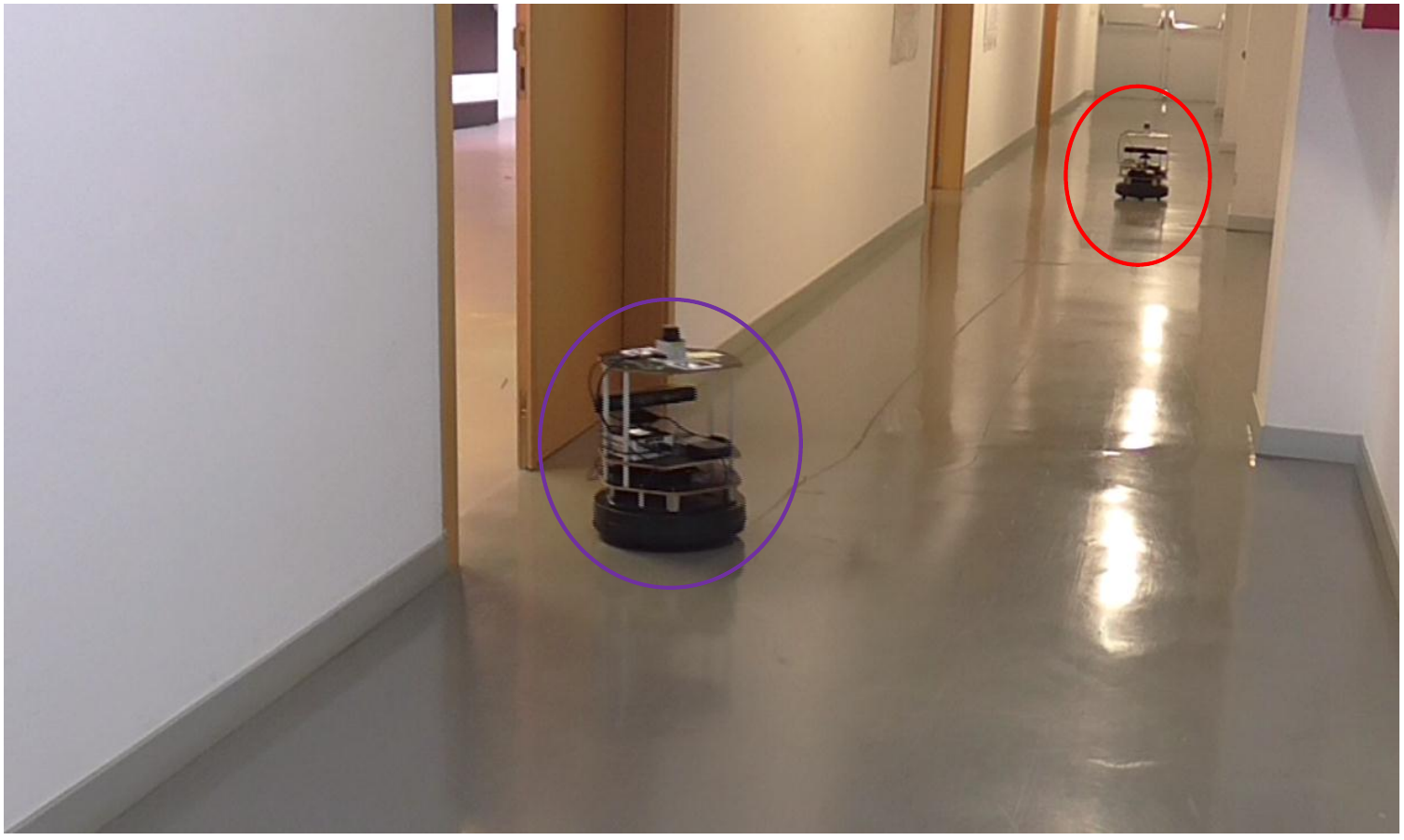}
                \label{Subfig:Snapshot21}
                }
                \hfil
        \subfloat[Small room and a part of the main room at $k=500$.]{
                \includegraphics[width=0.3\textwidth]{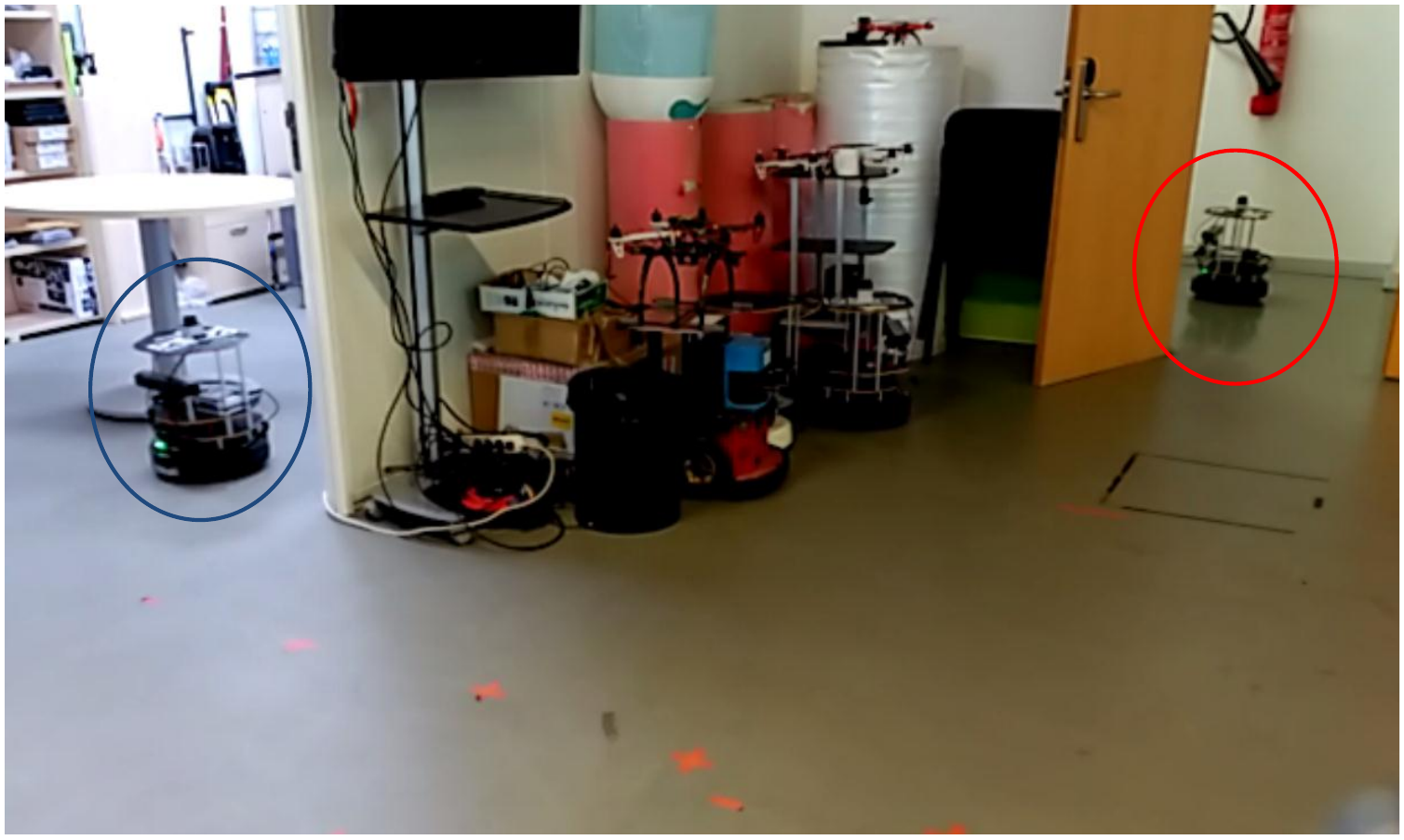}
                \label{Subfig:Snapshot22}
                }
                \hfil
        \subfloat[Coverage map and paths at $k=500$.]{
                \includegraphics[width=0.3\textwidth]{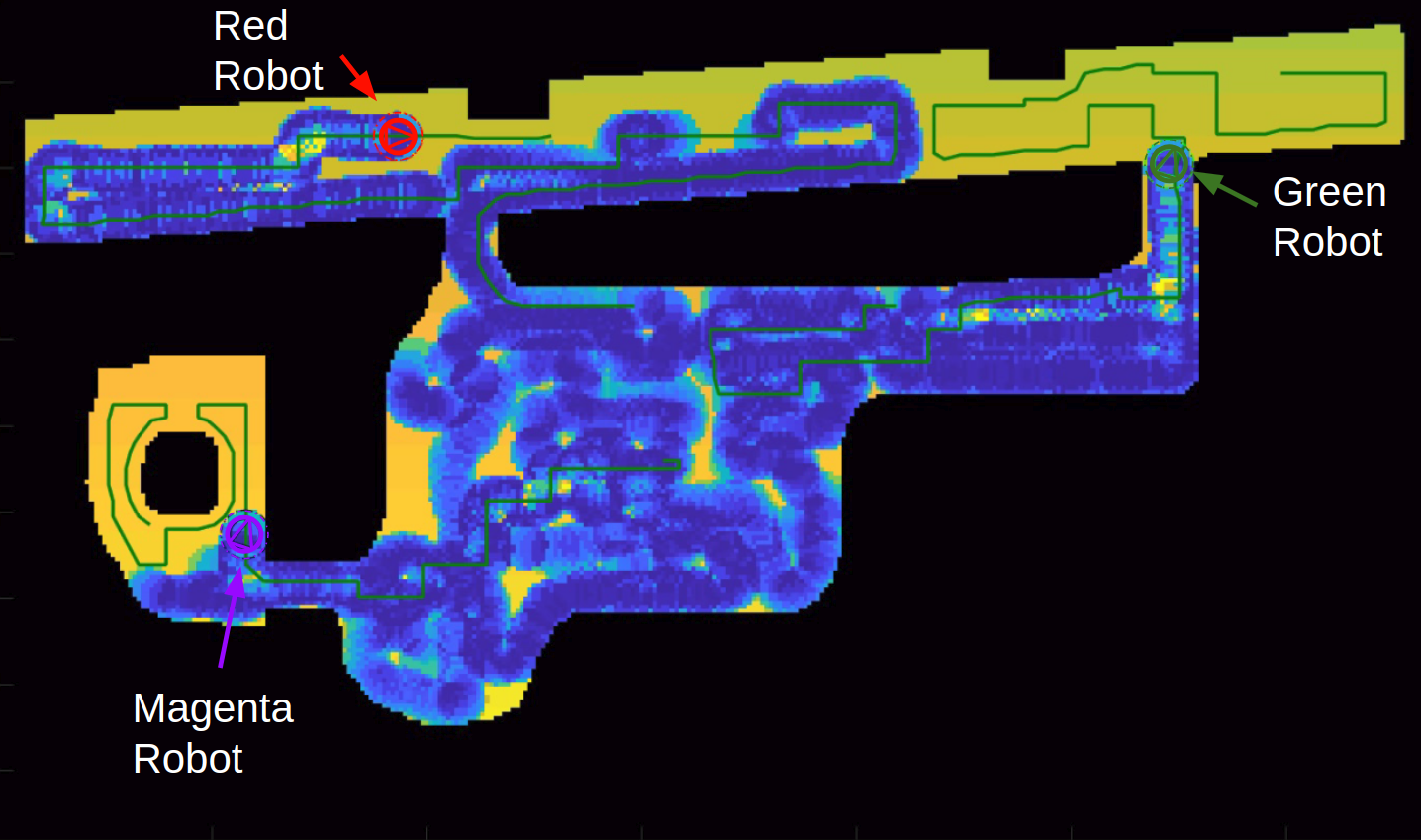}
                \label{Subfig:Snapshot2Matlab}
                }
    \caption{Snapshots of one of the experiments. The colored circles show the identity of each robot.}
    \label{Fig:Snapshots}
\end{figure*}

In Fig.~\ref{Fig:Snapshots} we show some snapshots of experiment B at three different iterations, $k=0$, $k=100$ and $k=500$. In particular, Fig.~\ref{Subfig:Snapshot0} shows the initial position of the robots while Fig.~\ref{Subfig:Snapshot1} shows the three robots in the main room covering their respective partitions after start-up. 
Even when the robot associated to the red partition starts outside of its partition, the planner drives it towards it improving the coverage of the other partition while going there. This coverage is taken into account by the planner in the magenta partition.
They start sweeping the environment from bottom to top, according to the importance of the points, as shown in Fig.~\ref{Subfig:Snapshot1Matlab}. After 500 iterations, the robots have covered almost the entire environment for the first time. This can be seen in Fig.~\ref{Subfig:Snapshot2Matlab}. At that point, the blue robot is inside the small room covering the surroundings of the table (Fig.~\ref{Subfig:Snapshot21}) and the other two are sweeping the corridor (Fig.~\ref{Subfig:Snapshot22}). It is worth highlighting how the coverage of the central and lower part of the main room had already decayed at $k=500$, since the were the first covered areas.

Eventually, we analyze the results from experiments A and B. It can be seen that in both cases they are very similar to the simulation results, which validates our approach for a real system in a complex environment.

\begin{figure}[!ht]
        \centering
        \subfloat[Mean coverage error of the environment (solid line) and its standard deviation (dashed line).]{
                \includegraphics[width=0.4\textwidth]{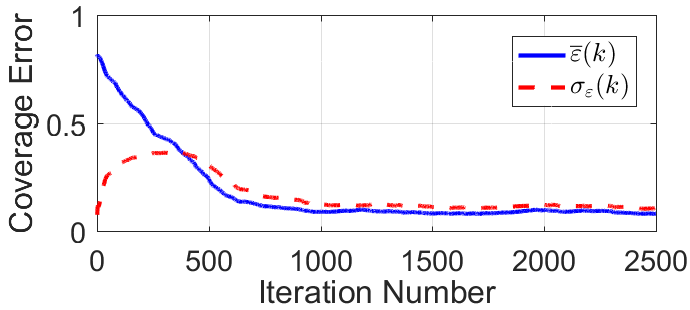}
                \label{Subfig:CovErrorExp2}
                }
                
        \subfloat[Mean coverage error of each partition (solid lines) and its standard deviation (dashed lines).]{
                \includegraphics[width=0.4\textwidth]{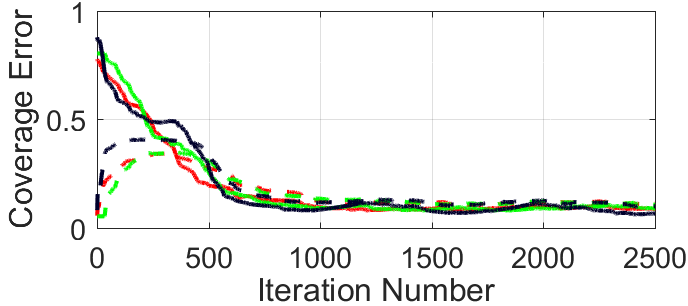}
                \label{Subfig:CovErrorPartExp2}
                }
                
        \subfloat[Mean coverage level of the environment (solid blue line), its standard deviation (dashed red line), mean coverage objective (dash-dotted blue line) and mean objective weighted by the workload (dash-dotted green line).]{
                \includegraphics[width=0.4\textwidth]{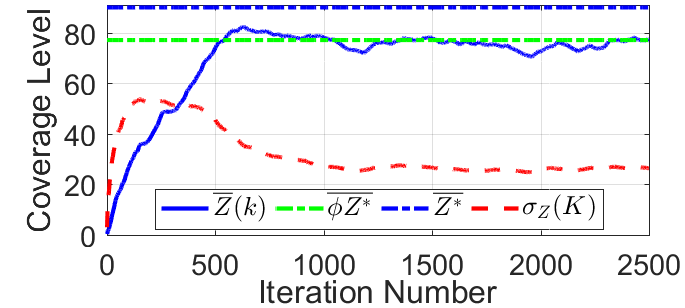}
                \label{Subfig:CovLevelExp2}
                }
                
        \subfloat[Paths of the robots until $k=1500$.]{
                \includegraphics[width=0.4\textwidth]{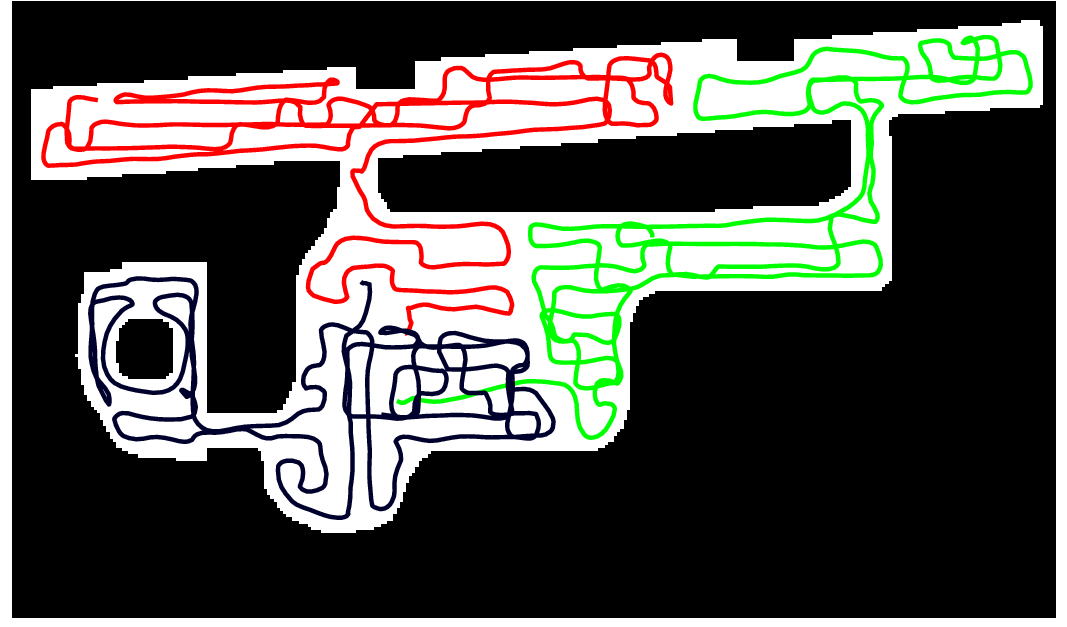}
                \label{Subfig:PathsExp2}
                }
    \caption{Results from experiment A.}
    \label{Fig:Exp2}
\end{figure}

In experiment A the coverage error in the entire environment (Fig.~\ref{Subfig:CovErrorExp2}) and in each partition (Fig.~\ref{Subfig:CovErrorPartExp2}) decreased rapidly and was maintained at a low value. The mean coverage level was maintained very close to the desired level, with almost the same value as the mean of the objective weighted by the work required at each point (Fig.~\ref{Subfig:CovLevelExp2}). This demonstrated that, although the robots could not maintain the desired level all the time because of the nature of the problem, they were able to maintain the most important points closer to their desired value. Fig.~\ref{Subfig:PathsExp2} depicts the real paths that the robots followed to cover the entire environment the first time.

\begin{figure}[!ht]
        \centering
        \subfloat[Mean coverage error of the environment (solid line) and its standard deviation (dashed line).]{
                \includegraphics[width=0.4\textwidth]{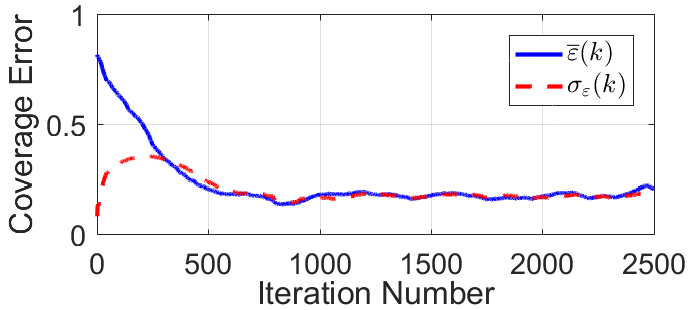}
                \label{Subfig:CovErrorExp3}
                }
                
        \subfloat[Mean coverage error of each partition (solid lines) and its standard deviation (dashed lines).]{
                \includegraphics[width=0.4\textwidth]{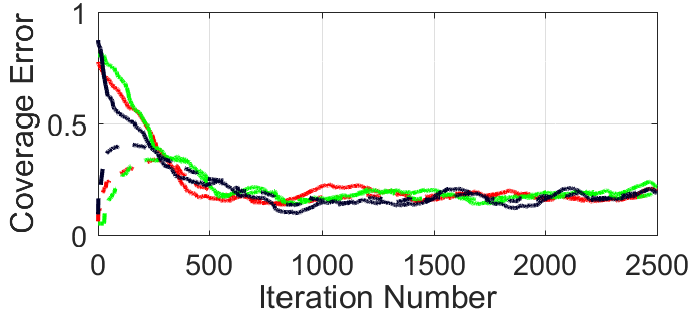}
                \label{Subfig:CovErrorPartExp3}
                }
                
        \subfloat[Mean coverage level of the environment (solid blue line), its standard deviation (dashed red line), mean coverage objective (dash-dotted blue line) and mean objective weighted by the workload (dash-dotted green line).]{
                \includegraphics[width=0.4\textwidth]{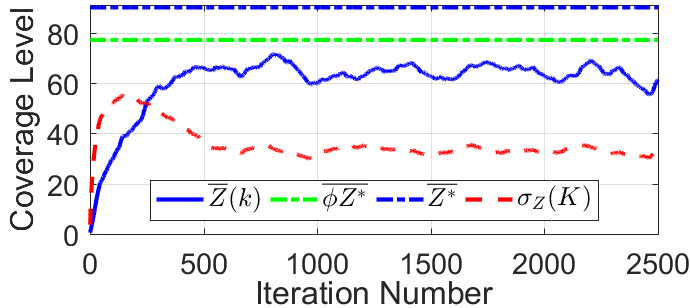}
                \label{Subfig:CovLevelExp3}
                }
                
        \subfloat[Paths of the robots until $k=1500$.]{
                \includegraphics[width=0.4\textwidth]{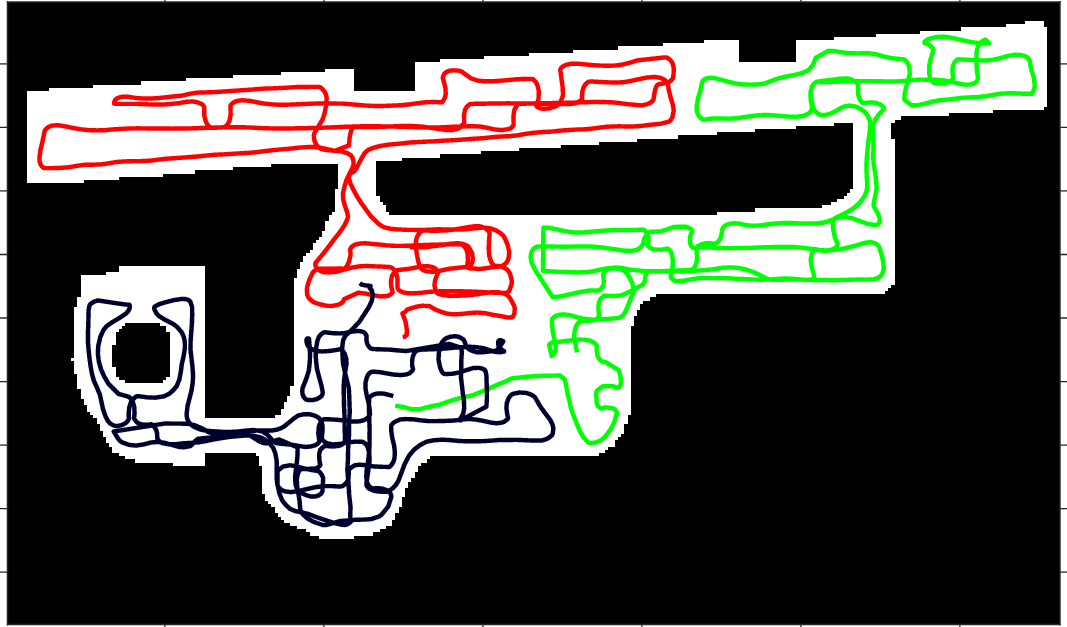}
                \label{Subfig:PathsExp3}
                }
    \caption{Results from experiment B.}
    \label{Fig:Exp3}
\end{figure}

In experiment B we decreased the value of the decay to simulate a faster deterioration of the coverage, i.e., to entail the robots a bigger workload and, therefore, forcing them to visit the important points with more frequency. This is equivalent to reducing the team size and deploying less robots than required by the environment.
For these reasons, the results show in this case a smaller reduction of the coverage error (Fig.~\ref{Subfig:CovErrorExp3}-~\ref{Subfig:CovErrorPartExp3})), a lower value of the coverage level and a slightly bigger standard deviation (Fig.~\ref{Subfig:CovLevelExp3}). The paths followed by the robots demonstrate that they covered several times the lower parts of their partitions, i.e., the most important ones, before covering their upper parts completely.
These experiments support the adaptability of our approach to environments with non-uniform workloads and, in particular experiment B, that the robots are able to prioritize the most important areas even if their capabilities are lower than the work required by the environment.

\section{Conclusions}
\label{sec:conclusions}

In this paper we have proposed a partitioning approach based on power diagrams along with a graph-based coverage planning strategy to solve the persistent coverage problem.
We have introduced in the first place an algorithm to find the power weights that correspond an equitable partition of the environment.
This strategy can be applied to any kind of non-convex environment thanks to the use of the geodesic distance in the partitioning. 
We have also presented two extensions for this strategy. The first extension drives the generator of each partition to the centroid of the connected component with the highest workload. The objective of this strategy is to reduce the disconnections of the partitions to the extent possible.
The second extension takes into account the different coverage capabilities of
each robot. 
In the second place, we have presented an online planning algorithm that allows each robot to locally find the best path in terms of the current coverage error.
This algorithm is based on the construction of a path graph that covers the entire partition of the robot with sweep-like paths. We assign the coverage errors of the head nodes to the edge weights and, therefore, the problem reduces to finding the optimal path through a graph, that can be achieved efficiently with state-of-the-art methods.
Finally, we include simulation and experimental results to support our contributions and show how our solution can be applied to a real world scenario.

\begin{funding}
This work was partially supported by projects RTC-2014-1847-6 and RTC-2017-5965-6 of Retos-Colaboraci\'on, DPI2015-69376-R and PGC2018-098719-B-I00 (MCIU/AEI/FEDER, UE), DGA-T45\_17R/FSA, CUD2017-18 and DGA Scholarship C076/2014, partially funded by European Social Fund.
\end{funding}

\bibliographystyle{SageH}

\end{document}